\documentclass{bmvc2k}

\usepackage{amsmath}
\usepackage{float}
\usepackage{amssymb}
\usepackage{relsize}
\usepackage{multirow}
\usepackage[flushleft]{threeparttable}
\usepackage{minibox}
\usepackage{wrapfig,booktabs}

\title{Hybrid Deep Network for Anomaly Detection}

\addauthor{Trong-Nguyen Nguyen}{nguyetn@iro.umontreal.ca}{1}
\addauthor{Jean Meunier}{meunier@iro.umontreal.ca}{1}

\addinstitution{Image Processing Laboratory\\
 DIRO, University of Montreal\\
 Montreal, QC, Canada
}

\runninghead{Nguyen, Meunier}{Hybrid Deep Network for Anomaly Detection}

\begin{document}

\maketitle

\begin{abstract}
In this paper, we propose a deep convolutional neural network (CNN) for anomaly detection in surveillance videos. The model is adapted from a typical auto-encoder working on video patches under the perspective of sparse combination learning. Our CNN focuses on (unsupervisedly) learning common characteristics of normal events with the emphasis of their spatial locations (by supervised losses). To our knowledge, this is the first work that directly adapts the patch position as the target of a classification sub-network. The model is capable to provide a score of anomaly assessment for each video frame. Our experiments were performed on 4 benchmark datasets with various anomalous events and the obtained results were competitive with state-of-the-art studies.
\end{abstract}

%-------------------------------------------------------------------------
\section{Introduction}
\label{sec:intro}
Anomaly detection in surveillance videos is currently getting attention for the development of automatic vision systems. Since anomalous events rarely happen in a long video, a method that can determine the potential of such events at frame-level is necessary.

Many approaches have been proposed to deal with this problem by processing either entire video frames~\cite{Li2014Anomaly,Zhang2016Video,Hasan2016Learning,Wen2018Future} or small image patches~\cite{Kim2009Observe,Zhao2011Online,Cheng2015Video,Xu2015Learning,Luo2017A}. Our work is inspired by the study~\cite{Lu2013Abnormal} belonging to the latter category. Specifically,~\citet{Lu2013Abnormal} partition the image plane into non-overlapping regions and then concatenate a number of successive frames to obtain spatio-temporal cuboids. The sparse combination learning is applied on such cuboids of the same spatial position. In other words, the anomaly assessment is performed independently on each region of input frames.% (see \colorbox[rgb]{1,1,0}{Fig.XXX}).

Following the spectacular development of deep CNNs in various vision problems such as image classification~\cite{Krizhevsky2012ImageNet,He2016Deep} and object detection~\cite{Shaoqing2015Faster,He2017Mask}, we attempt to replace the sparse combination learning applied for each spatial region by a convolutional auto-encoder (AE). The former algorithm focuses on representing an unknown input as a weighted combination of training samples while the latter one emphasizes their common characteristics. A trivial replacement leads to the use of a number of AEs where each one is assigned to a specific image patch position. Such ensemble of AEs requires huge resources such as memory for execution and storage capacity for storing the model parameters. In order to avoid these problems, we adaptively combine all AEs into a single one by adding a classification sub-network that forces the emphasized features to contain the information related to their corresponding spatial location. To the best of our knowledge, this is the first study adapting image patch positions as the classes in a supervised task.

The main contributions of this paper are summarized as: (1) integrating a classification sub-network into a convolutional AE model as a constraint of feature learning, (2) demonstrating that the combination of unsupervised and supervised objective functions provides better results than using only supervised learning for our anomaly detection, (3) a scheme combining scores obtained from different network components for improving the final anomaly assessment score, (4) experimental results on 4 benchmark datasets validating the competitive performance of the proposed model compared with state-of-the-art methods, and (5) a discussion regarding to the impact of the decoder in our hybrid network.

\section{Related Works}
One significant challenge of anomaly detection in surveillance videos is the diversity of anomalous events. In order to simplify this task, some studies (such as~\cite{Medioni2001Event,Zhang2009Learning,Robert2015Spatiotemporal}) focus on one of common basic factors: motion trajectory. By estimating that feature in videos, the original task becomes a typical problem of novelty detection. Such approaches can be easy to implement and have a fast execution speed. However, considering only motion trajectory does not cover the high diversity of possible anomalies. Besides, a failure in object localization and/or tracking would significantly reduce the efficiency of anomaly detection.

Another popular method is the use of sparse coding where the training data (containing only normal events) are a collection of small pieces (e.g. image patches, 3D cuboids and/or their features) and a new input is then represented as a combination of training patterns. A normal event is expected to provide a reconstructed combination result with small error and vice versa. These methodologies have reported to give efficient performance in many studies~\cite{Kim2009Observe,Zhao2011Online,Lu2013Abnormal}. A drawback of most approaches with this perspective is the high computational cost during the stage of searching sparse combination.

Recent studies employ deep neural networks with various architectures to perform image reconstruction~\cite{Hasan2016Learning} and/or translation~\cite{Ravanbakhsh2017Abnormal,Ravanbakhsh2019Training} and use the estimated loss as the measure of normality in video frames. These methods can be applied on the whole frames~\cite{Wen2018Future,Hasan2016Learning} or image patches~\cite{Xu2017Detecting,Sabokrou2018Adversarially}. Our model also falls into this category but integrates a classification sub-network to adapt the perspective of the work~\cite{Lu2013Abnormal} belonging to the previous one. Our hybrid network is expected to take advantages from both categories.

\section{Hybrid Deep Network}
As mentioned in the previous section, the hybrid model consists of two principal streams. The first one is a typical AE focusing on emphasizing common features of normal events at patch-level. The second stream performs a classification on the extracted characteristics and can be considered as a constraint for the former stream. Similarly to~\cite{Lu2013Abnormal}, we work on 3D cuboid of small spatial resolution of frame concatenation, in which these frames are resized to $160\times120$ and are gray-scale. However, we consider only 3 successive frames instead of 5 as in that study for faster execution without significant performance loss. The input of our model is thus a cuboid of size $10\times10\times3$. An overall architecture of the hybrid network is presented in Fig.~\ref{fig:architecture}(a). More details can be found in the supplementary material.
\begin{figure}[t]
%\center
\begin{picture}(520,138)
\put(0,0){\includegraphics[width=0.65\textwidth]{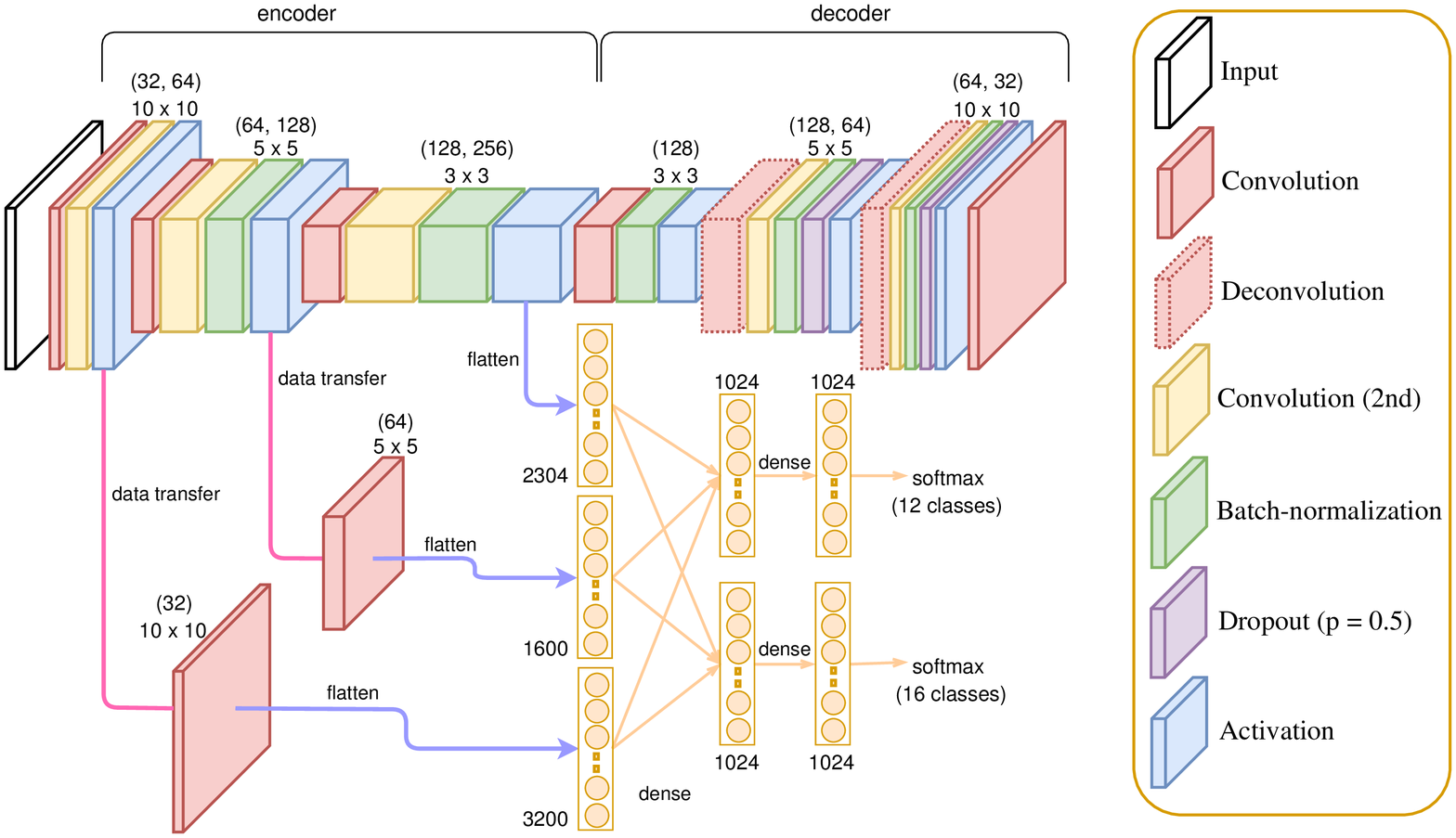}}
\put(244,0){\line(0,1){135}}
\put(248,30){\includegraphics[width=0.32\textwidth]{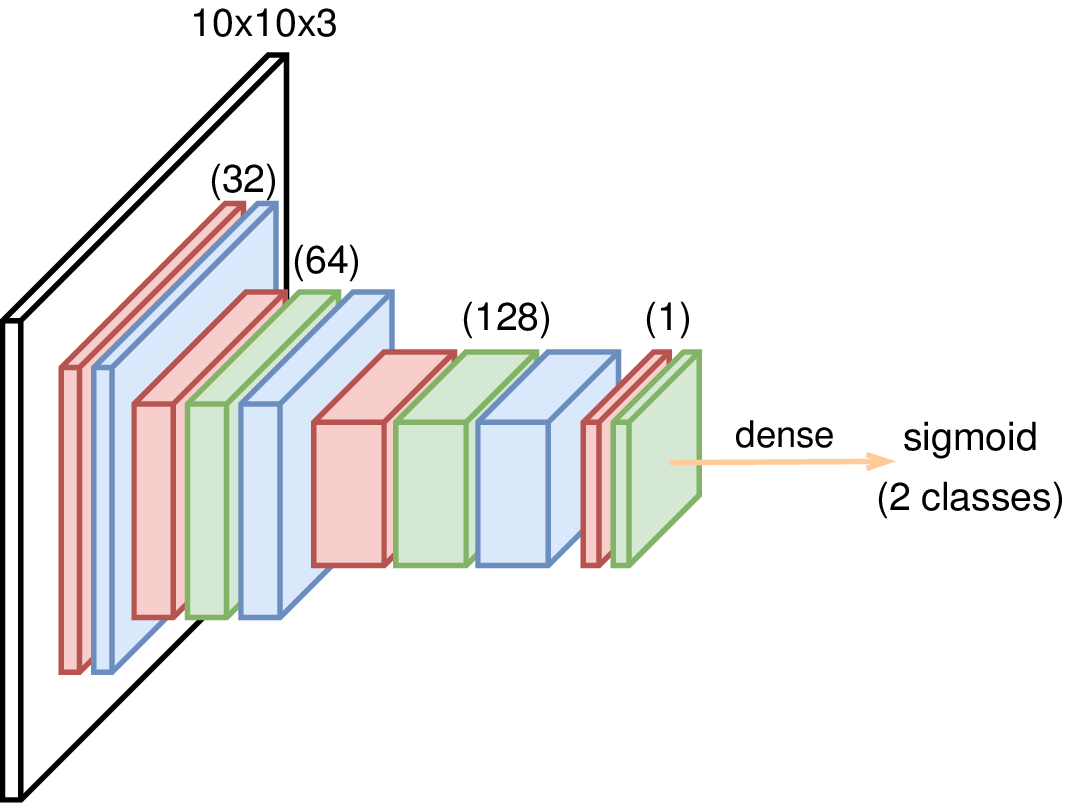}}
\put(160,0){(a)}\put(300,0){(b)}
\end{picture}
\caption{(a) Our hybrid network where input is a cuboid of size $10\times10\times3$.~\textit{Data transfer} indicates that the output of a layer is used by multiple next ones. The spatial dimensions of feature maps in each block are provided together with their number of channels (in parentheses). Notice that there are usually 2 numbers of channels because of the second convolutional layer (colorized as yellow). (b) The discriminator used for adversarial training in Section~\ref{sec:adv}. Best viewed in color.}
\label{fig:architecture}
\end{figure}

\subsection{Convolutional auto-encoder for learning common features}
The AE in our hybrid network aims to learn common local features of normal events. It works as a reconstruction model with a bottleneck structure. The temporal factor is embedded into this learning by concatenating consecutive frames. Some studies (e.g.~\cite{Xu2015Learning,Wen2018Future}) instead employ optical flow to represent this information. Such estimators may produce noisy results and/or require a training stage, e.g. FlowNet~\cite{Dosovitskiy2015FlowNet} and its upgraded version~\cite{Eddy2017Flownet2}.

As visualized in Fig.~\ref{fig:architecture}(a), there are two main components in our AE. The encoder is formed as a stack of layer blocks that reduce the resolution of feature maps, in which each block consists of two convolutional (Conv) layers, batch normalization, and activation. The first Conv layer converts the spatial dimensions of its input and the second one then transforms the features to a specific number of channels. Although these two operations can be performed with only one Conv layer, recent studies demonstrated good efficiencies when using this trick~\cite{Wen2018Future}. The use of the first Conv layer can be considered as an alternative to the pooling operation. In the decoder, this layer is replaced by deconvolution to upscale the spatial dimensions. Besides, the dropout (with probability set as 0.5) is integrated right before the activation layer. Notice that there is no batch normalization in the first block of the encoder and we use different activations for the two components (leakyReLU~\cite{Maas2013Rectifier} for the encoder and ReLU~\cite{Nair2010Rectified} for the decoder) as suggested in~\cite{Isola2017Image} although the skip connection is not employed in our model. The last layer of the decoder is a Conv layer that transforms the reconstructed features into the original size of $10\times10\times3$.

\subsection{Classification sub-network}\label{sec:sub-network}
Besides learning common local features in cuboids of normal events by the AE, we also add a constraint to the encoder so that the extracted information reveals spatial locations. To do that, the outputs of encoder blocks are vectorized and then concatenated to give a feature vector of 7104 elements. Notice that a convolution of $1\times1$ filters is performed before the vectorization to reduce the number of channels except for the latent variables [see Fig.~\ref{fig:architecture}(a)].

Since each frame of size $160\times120$ is partitioned into non-overlapping $10\times10$ patches, there are a total of $16\times12=192$ possible locations for an input cuboid. A classification problem with 192 classes on such small cuboid may be too complicated. In order to simplify this task, each patch position is represented by its spatial dimensions. Therefore, the classification is performed on two branches predicting the horizontal and vertical indices of image patches. The total number of classes is hence reduced to $16+12=28$. In addition, such architecture also allows the model to learn common features of cuboids belonging to the same row and the same column. This sub-network is a stack of dense layers with ReLU activation followed by a dropout, and the final outputs are provided by a softmax layer.

\subsection{Model optimization}
The optimization of our hybrid network is performed according to the objective functions of its partial components: reconstruction loss for the AE, classification loss for the location prediction sub-network, and adversarial training as an enhancement for reconstruction.

\subsubsection{Reconstruction loss}
This loss is defined as the difference between an input cuboid and its reconstructed result. In this work, we use the typical $l_2$ distance to measure this factor. A drawback of $l_2$ distance is the blur occurring in the obtained cuboid. To reduce this effect, an additional constraint on image spatial gradients is added into the objective function. This regularization improved the results provided by the decoder as reported in~\cite{Mathieu2015Deep,Wen2018Future}. Since the input cuboid is formed with the concatenation of successive image patches in temporal order, we also consider the gradient along this axis as a constraint related to the local motion. Given an input cuboid $\mathbf{c}$ and a convolutional auto-encoder $\mathcal{M}$, the final reconstruction loss is a combination of mentioned losses in an unsupervised learning way as
\begin{equation}
	\mathcal{L}_R(\mathbf{c},\mathcal{M})=\lambda_{l_2}\big\|\mathbf{c}-\mathcal{M}(\mathbf{c})\big\|^2_2 + \lambda_\nabla\sum_{d\in\{x,y,t\}}\big\||\nabla_d(\mathbf{c})|-|\nabla_d(\mathcal{M}(\mathbf{c}))|\big\|_1
	\label{eq:recon_loss}
\end{equation}
where $\lambda$s are the positive weights controlling the contribution of partial losses while $x$, $y$ and $t$ respectively indicate the cuboid's spatial and temporal dimensions. Notice that the gradient constraint is used in our objective function according to its efficiency in deep networks for reconstruction and frame prediction while 3D gradients are also employed in~\cite{Lu2013Abnormal} for feature description.

\subsubsection{Classification loss}
As mentioned above, the classification sub-network is to force learned common features to characterize cuboid spatial locations. The corresponding loss is computed as the summation of two dimensional predictions. Let $\mathcal{P}_x(\mathbf{c},l_x)$ and $\mathcal{P}_y(\mathbf{c},l_y)$ be the posterior probabilities of class labels $l_x$ and $l_y$ for the two classifiers given a cuboid $\mathbf{c}$ at position $(\mathbf{c}_x, \mathbf{c}_y)$, our classification loss is defined as the summation of cross-entropies as
\begin{equation}
	\mathcal{L}_C(\mathbf{c})=-\sum_{l_x}\mathrm{I}(l_x,\mathbf{c}_x)\log\big[\mathcal{P}_x(\mathbf{c},l_x)\big]-\sum_{l_y}\mathrm{I}(l_y,\mathbf{c}_y)\log\big[\mathcal{P}_y(\mathbf{c},l_y)]
	\label{eq:clf_loss}
\end{equation}
where $\mathrm{I}(\cdot,\cdot)$ returns 1 if the two operands are identical and 0 otherwise. There are a total of 16 labels for $l_x$ and 12 for $l_y$ corresponding to the classes along the horizontal and vertical dimensions since the resolution of input frames is $160\times120$ with patch size of $10\times10$.

\subsubsection{Adversarial training}\label{sec:adv}
Besides the two mentioned losses, another term is added into the objective function as an attempt to enhance the quality of reconstructed cuboids. Specifically, we design an additional discriminative model to form a generative adversarial network (GAN)~\cite{Goodfellow2014Generative}. A GAN includes a generator $\mathcal{G}$ that outputs desired data and a discriminator $\mathcal{D}$ which is a binary classifier attempting to distinguish such data with real patterns in training set. In other words, the two GAN components focus on opposite purposes in a game theory manner: $\mathcal{G}$ tries to fool $\mathcal{D}$ by generating outputs that are similar to training samples while $\mathcal{D}$ attempts to classify such outputs as fake data. The use of GAN architecture as a regularization component has achieved good performance for video frame prediction~\cite{Mathieu2015Deep,Wen2018Future} and image translation~\cite{Isola2017Image}.

In this work, the hybrid model in Fig.~\ref{fig:architecture}(a) plays the role of $\mathcal{G}$ where the generated data are cuboids reconstructed by the AE, and the model $\mathcal{D}$ (with structure provided in Fig.~\ref{fig:architecture}(b) and supplementary material) performs a binary classification with sigmoid activation. Given input cuboids $\mathbf{c}$ sampled from training data and the auto-encoder $\mathcal{M}$, the two opposite purposes of $\mathcal{G}$ and $\mathcal{D}$ are represented by the two following losses:
\begin{equation}
	\mathcal{L}_\mathcal{G}(\mathbf{c},\mathcal{M})=-\lambda_\mathcal{G}\log\mathcal{D}(\mathcal{M}(\mathbf{c}))+\lambda_R\mathcal{L}_R(\mathbf{c},\mathcal{M})+\lambda_C\mathcal{L}_C(\mathbf{c})
	\label{eq:G}
\end{equation}
\begin{equation}
	\mathcal{L}_\mathcal{D}(\mathbf{c},\mathcal{M})=-\frac{1}{2}\log\mathcal{D}(\mathbf{c})-\frac{1}{2}\log\big[1-\mathcal{D}(\mathbf{\mathcal{M}(\mathbf{c})})\big]
	\label{eq:D}
\end{equation}
where $\lambda_\mathcal{G}$, $\lambda_R$, $\lambda_C$ are respectively the positive weights controlling the contribution of adversarial, reconstruction and classification losses in $\mathcal{G}$. The optimization is performed by alternatively minimizing the two objective functions $\mathcal{L}_\mathcal{G}$ and $\mathcal{L}_\mathcal{D}$. Note that the discriminator $\mathcal{D}$ is only employed during the optimization, it does not play any role in the detection stage.

\subsection{Frame-level anomaly detection}
Once the optimization is completed, our hybrid network is capable to reconstruct an input cuboid $\mathbf{c}$ via the auto-encoder $\mathcal{M}$ as well as to provide softmax outputs $\mathcal{F}_x(\mathbf{c})$ and $\mathcal{F}_y(\mathbf{c})$ for spatial position classification. Denote the ground truth location of $\mathbf{c}$ by $(\mathbf{c}_x, \mathbf{c}_y)$, a cuboid can give three normality scores $\mathcal{S}_R(\mathbf{c})$, $\mathcal{S}_x(\mathbf{c})$ and $\mathcal{S}_y(\mathbf{c})$:
\begin{equation}
\mathcal{S}_R(\mathbf{c})=\mathrm{Max}\bigg(\big|\mathbf{c}-\mathcal{M}(\mathbf{c})\big|^\alpha\bigg);~\mathcal{S}_{d\in\{x,y\}}(\mathbf{c})=\mathrm{Mean}\bigg(\big|\mathcal{H}(\mathbf{c}_d)-\mathcal{F}_d(\mathbf{c})\big|^\beta\bigg)	
	\label{eq:score_cuboid}
\end{equation}
where $\mathcal{H}(\cdot)$ converts an input label into a one-hot vector, $\mathrm{Max}(\cdot)$ and $\mathrm{Mean}(\cdot)$ are functions respectively outputting the (scalar) maximum and average values.

With the three types of cuboid-level scores, we obtain 3 score maps of size $16\times12$ for each frame (actually a concatenation of it and two next consecutive frames). As an attempt to combine the three maps to provide an improved result, we estimate their weighted sum as the final normality score map:
\begin{equation}
	\mathcal{S}_{R,x,y}(\mathbf{c})=\mathlarger{‎‎\mathlarger{‎‎\sum}}_{k\in\{R,x,y\}}\underbrace{\bigg[1-\frac{1}{\|\mathbf{T}\|}‎‎\mathlarger{\sum}_{\mathbf{x}\in\mathbf{T}}\mathcal{S}_k(\mathbf{x})\bigg]}_\text{weight}\mathcal{S}_k(\mathbf{c})
	\label{eq:score_map}
\end{equation}
where $\mathcal{S}_k(\mathbf{c})$ denotes scores estimated on cuboid $\mathbf{c}$, and $\mathbf{T}$ indicates the collection of cuboids at the spatial location $(\mathbf{c}_x,\mathbf{c}_y)$ in the training set consisting of normal events. In summary, the weight of the cuboid at a specific position for each score type is assigned according to its efficiency on the training data (for which $\mathcal{S}_k$ should be small). Notice that Eq.~(\ref{eq:score_map}) can be easily implemented at frame-level where the multiplication is element-wise.

The frame-level normality measure, $s(\mathbf{f})$, is defined as the standard deviation (SD) of cuboid-level scores $\mathcal{S}_{R,x,y}(\mathbf{c})$ obtained from cuboids of the same frame. In related studies, this measure is estimated as the extrema in each score (or error) map~\cite{Ravanbakhsh2017Abnormal} or a distance between the ground truth and their model output~\cite{Hasan2016Learning,Wen2018Future}. We consider the SD according to the perspective that the cuboid-level scores in our map are expected to be distributed in a small range of values for normal frames and spread over a wider interval when the frame contains normal and abnormal image patches. In Section~\ref{sec:experiment}, we denote by $\mathcal{S}_{x,y}$ the combination of only $\mathcal{S}_x$ and $\mathcal{S}_y$, i.e. the weight of $\mathcal{S}_R(\mathbf{c})$ in Eq.~(\ref{eq:score_map}) is set to 0 in the calculation of $\mathcal{S}_{x,y}$. %We also use the notation $\mathcal{S}$ to indicate the score maps of size $16\times12$ in our experiments.

Similarly to related studies~\cite{Hasan2016Learning,Ravanbakhsh2017Abnormal,Wen2018Future}, our frame-level score $s(\mathbf{f})$ is normalized for each evaluated video of $n$ frames as
\begin{equation}
	\hat{s}(\mathbf{f}_i)=\frac{s(\mathbf{f}_i)}{\max\big[s(\mathbf{f}_1),...,s(\mathbf{f}_n)\big]}, 1\le i\le n
	\label{eq:norm_score}
\end{equation}
This score is expected to be high for frames with anomalous events and to be low otherwise.

\section{Experiments}\label{sec:experiment}
In this section, we provide evaluation results obtained on 4 benchmark datasets: CUHK Avenue~\cite{Lu2013Abnormal}, UCSD Ped2~\cite{Li2014Anomaly}, Belleview and Traffic-Train~\cite{Zaharescu2010Anomalous}. A comparison is also presented on related works that perform image patch processing. We assigned $\lambda_{l_2}=1$ and $\lambda_\nabla=3^{-1}$ in Eq.~(\ref{eq:recon_loss}) as a dimensional average of pixel gradient. In Eq.~(\ref{eq:G}), $\lambda_\mathcal{G}$ was set to 0.25, $\lambda_R$ and $\lambda_C$ were both 1. Regarding to the powers for cuboid-level normality measurement in Eq.~(\ref{eq:score_cuboid}), we empirically set $\alpha=1$ and $\beta=2$. In the training stage, the generator $\mathcal{G}$ was optimized using Adam algorithm~\cite{Adam2015A} and the typical gradient descent method was used for optimizing the discriminator $\mathcal{D}$. Their initial learning rates were respectively $2\times10^{-4}$ and $10^{-4}$. More details of experimental results can be found in the supplementary material.

\subsection{Datasets}
\paragraph{CUHK Avenue} This dataset was acquired in an avenue of a campus consisting of 16 clips (15328 frames) of only normal events for training and 21 clips (15324 frames) for evaluation. The anomalous events occurred in the test set include unusual behavior, movement (in speed and/or direction) and the appearance of vehicle.

\paragraph{UCSD Ped2} The UCSD dataset captures walkways with only pedestrians. There are two subsets Ped1 and Ped2 with different walkway orientations. Since they are similar, we considered only Ped2 of 4560 frames for the evaluation. The numbers of clips in the training and test sets are 16 and 12, respectively. The anomaly in this dataset is non-pedestrian objects.

\paragraph{Belleview} The data was acquired by a camera mounted at a high position overlooking a road intersection. The normality is defined as the movement of vehicles on the main street while the motion and/or appearance of ones on other roads is considered as anomaly. The training data consists of only 300 frames and there are 2618 frames in the test set.

\paragraph{Traffic-Train} This dataset was acquired from a surveillance camera on a moving train. The movement of passengers is considered as abnormal events. Differently from the above datasets, the Traffic-Train is more challenging due to the camera jitter and the sudden change of lighting. There are 800 frames for training the model and 4160 frames for evaluation.

\subsection{Evaluation metrics}
For Avenue and Ped2 datasets, we evaluated the network performance by the area under curve (AUC) of the receiver operating characteristic (ROC) curve estimated from the frame-level scores and the provided ground truth. The AUC metric was also used in most studies related to anomaly detection.

Regarding to Belleview and Train datasets, we employed the average precision (AP) estimated from the precision-recall (PR) curve for the assessment according to previous experiments on them ~\cite{Zaharescu2010Anomalous,Xu2015Learning}. Note that results reported in these related studies were obtained at the pixel-level instead of frame-level as in our experiments.

\subsection{Experimental results}\label{sec:results}
\begin{figure}[t]
\begin{center}
\begin{picture}(520,80)
\put(0,-15){\includegraphics[scale=0.38]{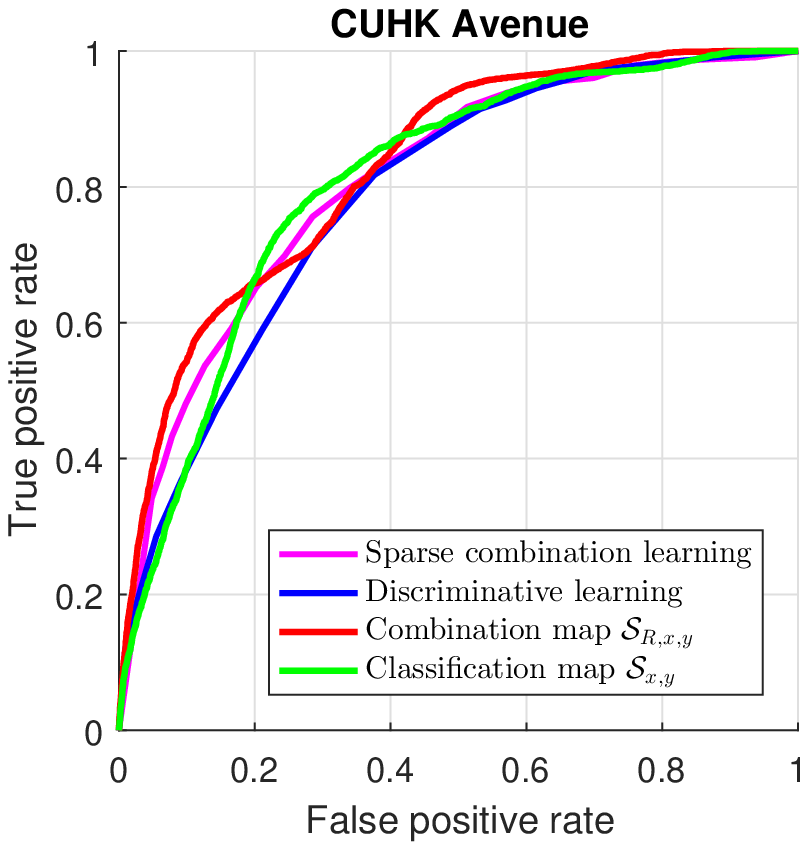}}
\put(92,-15){\includegraphics[scale=0.38]{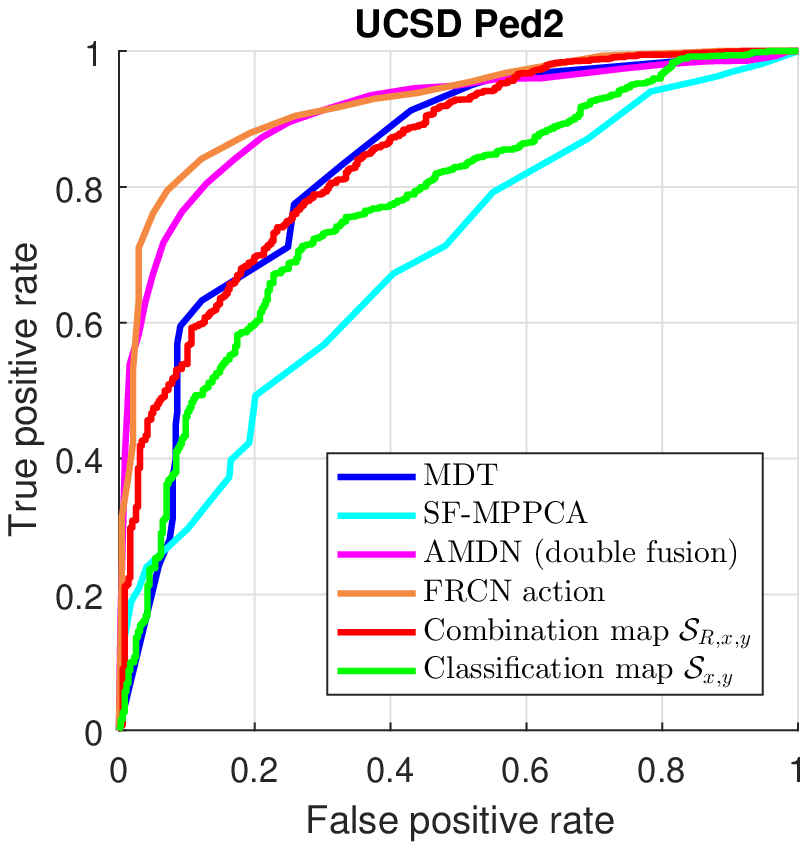}}
\put(184,-15){\includegraphics[scale=0.38]{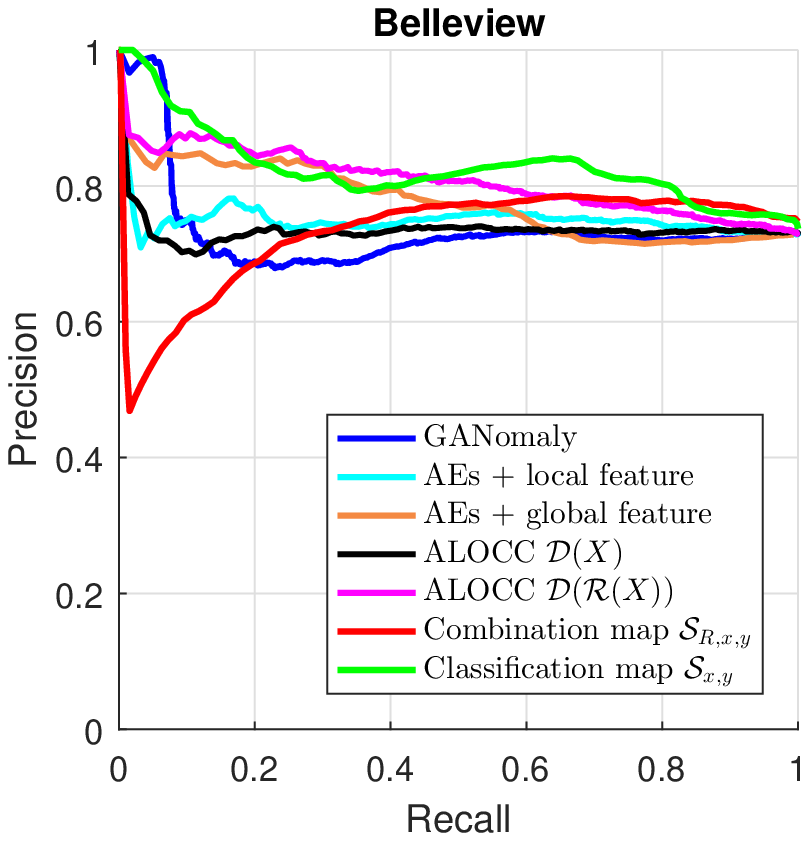}}
\put(276,-15){\includegraphics[scale=0.38]{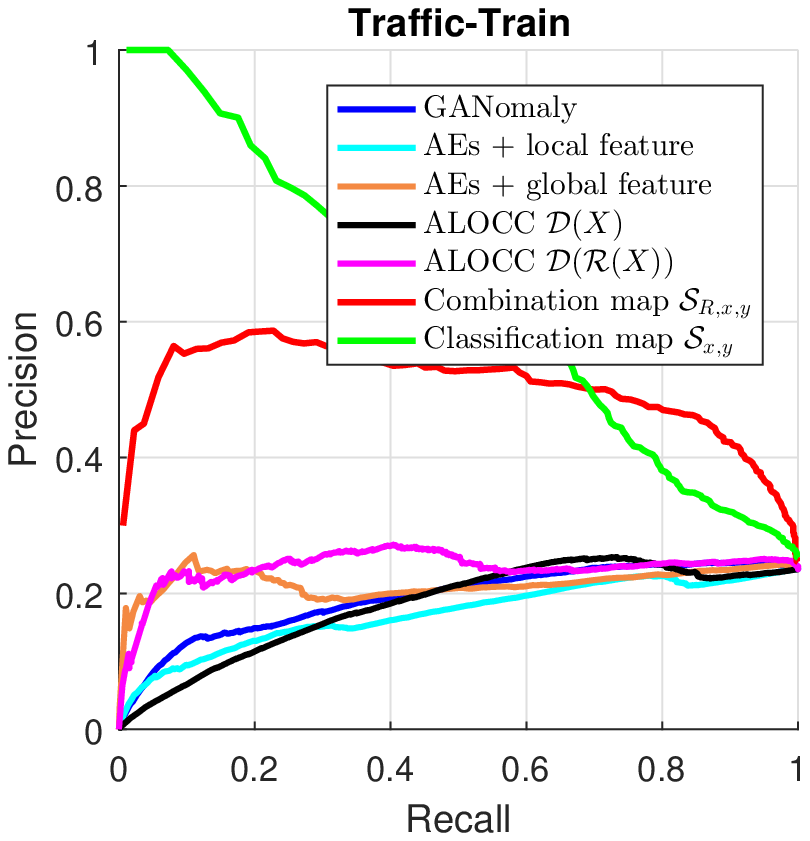}}
\end{picture}
\end{center}
\caption{Evaluation curves (frame-level) on 4 experimented datasets: ROC curve for Avenue and Ped2, PR curve for Belleview and Traffic-Train. Best viewed in color.}
\label{fig:graphs}
\end{figure}

A visualization of ROC and PR curves in our evaluation is given in Fig.~\ref{fig:graphs}. The curves corresponding to some related methods are also provided including MDT~\cite{Mahadevan2010Anomaly}, SF-MPPCA~\cite{Mahadevan2010Anomaly}, sparse combination learning~\cite{Lu2013Abnormal}, discriminative learning~\cite{Giorno2016A}, FRCN action~\cite{Hinami2017Joint}, AMDN (double fusion)~\cite{Xu2017Detecting}, GANomaly~\cite{Samet2018GANomaly}, AEs + local/global features~\cite{Narasimhan2018Dynamic} and ALOCC~\cite{Sabokrou2018Adversarially}.

\begin{table}[t]
\begin{center}
\footnotesize
\begin{tabular}{|l|l|c|c|c|}
\hline
\multirow{2}{*}{Method} & \multirow{2}{*}{Additional model} & \multirow{2}{*}{Optical flow} & \multicolumn{2}{c|}{AUC (\%)} \\ \cline{4-5} & & & Avenue & Ped2 \\
\hline\hline
MPPCA~\cite{Kim2009Observe} & - & - & - & 69.3 \\
SF-MPPCA~\cite{Mahadevan2010Anomaly} & - & - & - & 61.3 \\
MDT~\cite{Mahadevan2010Anomaly} & - & - & 81.8 & 82.9 \\
Sparse combination learning~\cite{Lu2013Abnormal} & - & - & 80.9 & - \\
Discriminative learning~\cite{Giorno2016A} & - & - & 78.3 & - \\
FRCN action~\cite{Hinami2017Joint} & AlexNet~\cite{Krizhevsky2012ImageNet} & - & - & \textbf{92.2} \\
Unmask (Conv5)~\cite{Ionescu2017Unmasking} & VGG-f~\cite{Chatfield2014Return} & - & 80.5 & 82.1 \\
Unmask (3D gradient)~\cite{Ionescu2017Unmasking} & VGG-f~\cite{Chatfield2014Return} & - & 80.1 & 81.3 \\
Unmask (late fusion)~\cite{Ionescu2017Unmasking} & VGG-f~\cite{Chatfield2014Return} & - & 80.6 & 82.2 \\
Stacked RNN~\cite{Luo2017A} & ConvNet & - & 81.7 & \textbf{92.2} \\
AMDN (early fusion)~\cite{Xu2017Detecting} & - & Yes & - & 81.5 \\
AMDN (double fusion)~\cite{Xu2017Detecting} & One-class SVM & Yes & - & 90.8 \\\hline
Reconstruction map $\mathcal{S}_R$ & - & - & 80.1 & 76.3 \\
%Classification map $\mathcal{S}_x$ & - & - & 82.0 & 75.1 \\
%Classification map $\mathcal{S}_y$ & - & - & 79.3 & 75.4 \\
Classification map $\mathcal{S}_{x,y}$ & - & - & 80.6 & 76.8 \\
Combination map $\mathcal{S}_{R,x,y}$ & - & - & \textbf{82.8} & 84.3 \\
%Our combination score & - & - & 82.8 & 84.3 \\
\hline
\end{tabular}
\end{center}
\caption{Comparison of AUCs provided from related studies and ours on CUHK Avenue and UCSD Ped2 datasets.}
\label{table:avenue_ped2}
\end{table}
\begin{table}[t]
\footnotesize
\begin{center}
\begin{threeparttable}
\begin{tabular}{|l|c|c|}
\hline
\multirow{2}{*}{Method} & \multicolumn{2}{c|}{Average precision (\%)} \\\cline{2-3}
& Belleview & Traffic-Train \\
\hline\hline
Sparse combination learning~\cite{Lu2013Abnormal}$^\ddag$ & 77.1 & 29.2 \\
GANomaly~\cite{Samet2018GANomaly} & 73.5 & 19.4 \\
AEs + local feature~\cite{Narasimhan2018Dynamic} & 74.8 & 17.1 \\
AEs + global feature~\cite{Narasimhan2018Dynamic} & 77.6 & 21.6 \\
ALOCC $\mathcal{D}(X)$~\cite{Sabokrou2018Adversarially} & 73.4 & 18.2 \\
ALOCC $\mathcal{D}(\mathcal{R}(X))$~\cite{Sabokrou2018Adversarially} & 80.5 & 23.7 \\\hline
Reconstruction map $\mathcal{S}_R$ & 68.6 & 40.0 \\
%Classification map $\mathcal{S}_x$ & \textbf{82.8} & 48.0 \\
%Classification map $\mathcal{S}_y$ & 82.7 & 45.8 \\
Classification map $\mathcal{S}_{x,y}$ & \textbf{82.7} & \textbf{64.4} \\
Combination map $\mathcal{S}_{R,x,y}$ & 73.1 & 50.5 \\
\hline
\end{tabular}
\begin{tablenotes}
  \item $^\ddag$\scriptsize{Evaluation on only $160\times120$ frames instead of multi-scale as in the paper.}
\end{tablenotes}
\end{threeparttable}
\end{center}
\caption{Evaluation results of experiments on Belleview and Traffic-Train datasets.}
\label{table:traffic}
\end{table}
AUCs obtained from our frame-level evaluation with score maps $\mathcal{S}_R$, $\mathcal{S}_{x,y}$ and $\mathcal{S}_{R,x,y}$ on the CUHK Avenue and UCSD Ped2 datasets are presented in Table~\ref{table:avenue_ped2}. A comparison with related studies that perform processing on local frame regions is also provided. This table shows that our method is better than others when working on the Avenue dataset. However, the gap between our performance and the best results on Ped2 dataset is nearly 8\%. This is possibly due to the large distance between the camera and scene in this dataset, a cuboid with size of $10\times10\times3$ thus does not contain enough useful details for our hybrid model (see Section~\ref{sec:Ped2dims}). It is also worth noting that the two methods~\cite{Hinami2017Joint,Luo2017A} require feature extractors pretrained on large datasets of object classification while our model was trained from scratch.

The evaluation results of the two remaining datasets are given in Table~\ref{table:traffic}. Two recent studies~\cite{Samet2018GANomaly,Sabokrou2018Adversarially} also employ GANs but in a different way compared with ours. Specifically, the generator AE in~\cite{Samet2018GANomaly} is followed by another encoder that provides a second latent variable supporting both optimization and inference stages while~\citet{Sabokrou2018Adversarially} directly use the discriminator for their normality measurement. Due to the low quality of video frames in Belleview dataset, our reconstruction score map does not work well since its AP is only 68.6\%. The classification sub-network, however, gives the best result with nearly 83\%. It is also observed that the combination of partial score maps according to Eq.~(\ref{eq:score_map}) may not enhance all the 3 inputs if the gap between reconstruction and classification efficiencies is significant. The removal of $\mathcal{S}_R$ for low-quality video hence seems an appropriate choice. In addition, the experimental results on the Traffic-Train dataset demonstrates that our hybrid network can deal with the sudden change of lighting as well as camera jitter in a better way compared with models of similar architectures in related works.

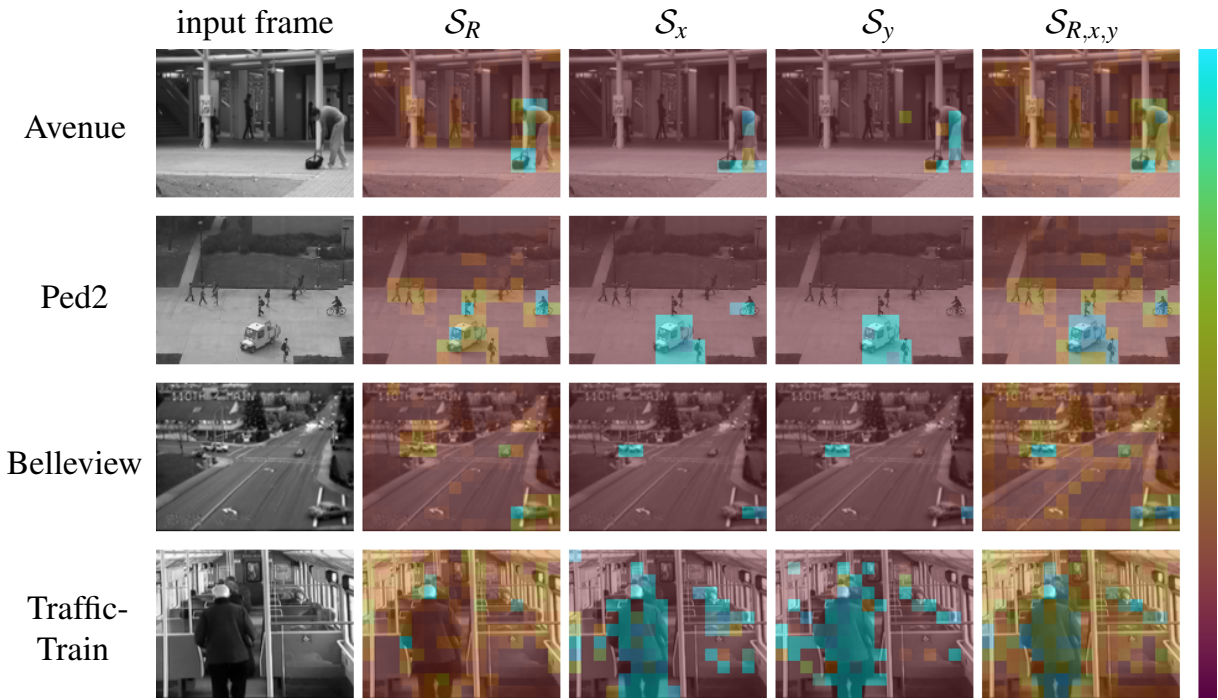
\begin{figure}[t]
\begin{center}
\begin{picture}(520,190)
\put(46,-15){\includegraphics[width=0.85\textwidth]{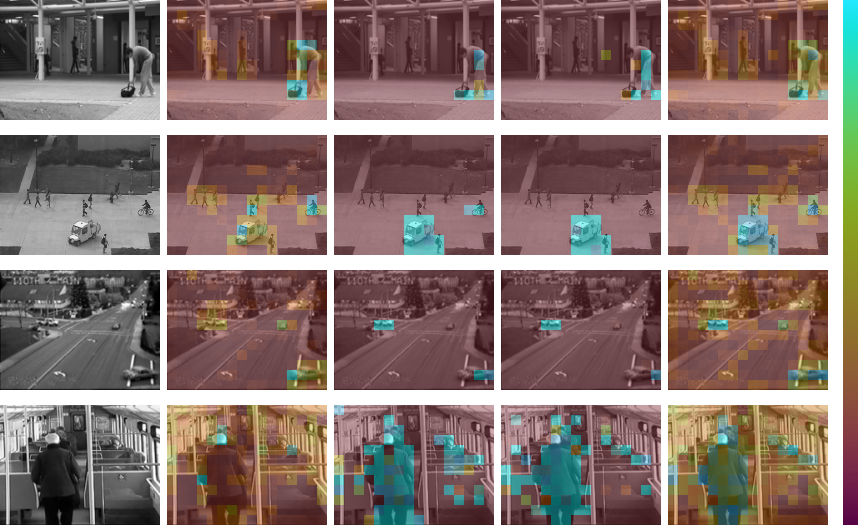}}
\put(52,181){input frame}\put(131,181){$\mathcal{S}_R$}\put(193,181){$\mathcal{S}_x$}
\put(254,181){$\mathcal{S}_y$}\put(308,181){$\mathcal{S}_{R,x,y}$}
\put(7,150){Avenue}\put(12,100){Ped2}\put(2,51){Belleview}\put(8,2){\minibox{Traffic-\\\hspace{0.1cm}Train}}
\end{picture}
\end{center}
\caption{Examples of our score maps (that superimpose the input frame) estimated on evaluation datasets. The color mapping of score is shown on the right. Best viewed in color.}
\label{fig:maps}
\end{figure}
An illustration of our score maps is presented in Fig.~\ref{fig:maps}. The input frame is superimposed to these individual maps to provide the visual correspondence between each cuboid spatial location and its responses. This figure shows that the reconstruction score map $\mathcal{S}_R$ is noisy but still emphasizes the saliency of anomalous events except for the Traffic-Train dataset due to the significant lighting change and camera jitter. On the contrary, the classification score maps $\mathcal{S}_x$ and $\mathcal{S}_y$ can better localize the anomaly but may provide isolated noises as we observed in the experiments. The combination of these maps, $\mathcal{S}_{R,x,y}$, can be considered as a smoothing operation that provides a balanced result.

The average time for forwarding a batch of 3072 cuboids of size $10\times10\times3$ through the network was 0.15 seconds using Python and TensorFlow on a computer with Intel i7-7700K, 16 GB memory, and GTX 1080. The model is thus expected to be appropriate for integrating into real-time systems.

\subsection{Impact of frame resolution on UCSD Ped2}\label{sec:Ped2dims}
\begin{wrapfigure}{l}{0.4\textwidth}
\vspace{-18pt}
\includegraphics[width=0.42\textwidth]{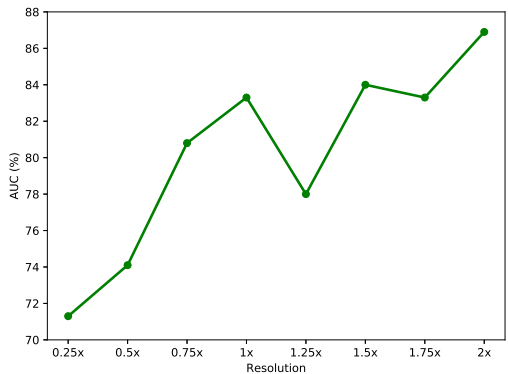}
\vspace{-23pt}
\caption{Frame-level AUCs obtained from anomaly detection on UCSD Ped2 with various frame resolutions. The base 1x is $160\times120$.}
\label{fig:Ped2dims}
\end{wrapfigure}
As mentioned in Section~\ref{sec:results}, a possible reason of unexpected detections on UCSD Ped2 is the lack of useful details in cuboids due to the large distance between the camera and scene. To verify this hypothesis, we performed the assessment on this dataset with various frame resolutions scaled from the base one $160\times120$. Since the patch size was still $10\times10$, only the number of classes changed, e.g. 32 and 24 classes for $320\times240$ frames. We trained all these models with only 40 epochs. This early stopping was performed since we focused on the potential of the proposed model architecture instead of attempting to find the best resolution for UCSD Ped2.

The AUCs obtained according to the normality score $\mathcal{S}_{R,x,y}$ are presented in Fig.~\ref{fig:Ped2dims}, in which increasing the frame resolution tends to improve the model ability. Therefore, the experimental results of our model on UCSD Ped2 can be expected to be better when upscaling the input frames to appropriate dimensions.

\subsection{Impact of adversarial training}\label{sec:impactGAN}
In order to assess the impact of the adversarial training described in Section~\ref{sec:adv}, we reimplemented the experiments using only the reconstruction and classification losses. Concretely, the discriminator $\mathcal{D}$ was not employed and the training was thus performed using the objective function $\mathcal{L}_\mathcal{G}$ in Eq.~(\ref{eq:G}) where $\lambda_\mathcal{G}$ was assigned to 0. The corresponding experimental results are presented in Table~\ref{table:impactGAN}. They show that using adversarial training improved the detection efficiency on most datasets except for the Traffic-Train. The reason is possibly the camera jitter which changed the structural texture of $10\times10$ patches. Since such variations (and the sudden change of lighting) were unpredictable, the discriminator $\mathcal{D}$ might encounter difficulties in distinguishing real cuboids from the decoder's output. Therefore, the use of adversarial training is unrecommended if the camera is unstable.

\begin{table}[H]
\scriptsize
\centering
\begin{tabular}{|l|ccc|ccc|ccc|ccc|}
\hline
Dataset & \multicolumn{3}{c|}{Avenue$^\dag$} & \multicolumn{3}{c|}{UCSD Ped2$^\dag$} & \multicolumn{3}{c|}{Belleview$^\ddag$} & \multicolumn{3}{c|}{Traffic-Train$^\ddag$} \\\hline
Normality score & $\mathcal{S}_R$ & $\mathcal{S}_{x,y}$ & $\mathcal{S}_{R,x,y}$ & $\mathcal{S}_R$ & $\mathcal{S}_{x,y}$ & $\mathcal{S}_{R,x,y}$ & $\mathcal{S}_R$ & $\mathcal{S}_{x,y}$ & $\mathcal{S}_{R,x,y}$ & $\mathcal{S}_R$ & $\mathcal{S}_{x,y}$ & $\mathcal{S}_{R,x,y}$ \\ \hline
w/ adv. training & 80.1 & 80.6 & \textbf{82.8} & 76.3 & 76.8 & \textbf{84.3} & 68.6 &\textbf{82.7} & 73.1 & 40.0 & 64.4 & 50.5 \\ %\hline
w/o adv. training & 80.1 & 80.3 & 82.5 & 74.3 & 74.0 & 77.4 & 68.3 & 81.7 & 72.5 & 42.9 & \textbf{67.4} & 54.9 \\ \hline
\end{tabular}
\caption{Comparison of anomaly detection ability between the proposed model and its modified version without adversarial training. $^\dag$AUC. $^\ddag$AP.}
\label{table:impactGAN}
\end{table}

\subsection{Impact of the decoder}\label{sec:decoder_test}
\begin{wraptable}{l}{0.33\textwidth}
\vspace{-12pt}
%\begin{center}
%\centering
\footnotesize
\begin{tabular}{|l|r|c|c|}
\hline
Dataset&metric&w/ & w/o \\
\hline\hline
Avenue & AUC& \textbf{80.6} & 80.3 \\
Ped2 & AUC&\textbf{76.8} & 73.6 \\
Belleview& AP & \textbf{82.7} & 81.6 \\
Train & AP&\textbf{64.4} & 63.3\\
\hline
\end{tabular}
%\end{center}
\caption{Evaluation results on $\mathcal{S}_{x,y}$. Notations \textit{w/} and \textit{w/o} are abbreviations of \textit{with} and \textit{without} the decoder, respectively.}
\label{table:without}
\vspace{-12pt}
\end{wraptable}
Typically, an AE is applied to (unsupervisedly) learn the underlying features of data while a classification deep network can implicitly perform the feature extraction. Therefore, our hybrid network can theoretically work without the decoder. To evaluate its effect, we re-performed the experiments after removing the decoder in Fig.~\ref{fig:architecture}(a). Besides, the adversarial training (see Section~\ref{sec:adv}) is also unnecessary since it focuses on enhancing the cuboids reconstructed from the AE. The network hence becomes a typical classification model with the objective function $\mathcal{L}_C$ in Eq.~(\ref{eq:clf_loss}) and provides only two score maps $\mathcal{S}_x$ and $\mathcal{S}_y$ for each input frame. Table~\ref{table:without} presents a comparison of the efficiency of $\mathcal{S}_{x,y}$ for the two networks (with and without decoder). It shows that the learning of common cuboid features via the bottleneck AE has a certain contribution for anomaly detection. Some learned convolutional filters in the two networks are shown in the supplementary material.

\vspace{-12pt}
\section{Conclusion}
In this paper, we propose a hybrid deep network combining both supervised and unsupervised learning perspectives for anomaly detection in surveillance videos. We designed a convolutional auto-encoder for learning common local appearance-motion features from small spatio-temporal cuboids. A classification sub-network is integrated to force such characteristics to be distinguishable for cuboids of different spatial locations. A novel anomaly score estimation/combination is presented and the impact of the decoder in our model is also discussed. The experiments on 4 benchmark datasets demonstrated the competitive performance of our hybrid network compared with state-of-the-art methodologies.

%\begin{figure}
%\begin{tabular}{ccc}
%\bmvaHangBox{\fbox{\parbox{2.7cm}{~\\[2.8mm]
%\rule{0pt}{1ex}\hspace{2.24mm}\includegraphics[width=2.33cm]{images/eg1_largeprint.png}\\[-0.1pt]}}}&
%\bmvaHangBox{\fbox{\includegraphics[width=2.8cm]{images/eg1_largeprint.png}}}&
%\bmvaHangBox{\fbox{\includegraphics[width=5.6cm]{images/eg1_2up.png}}}\\
%(a)&(b)&(c)
%\end{tabular}
%\caption{It is often a good idea for the first figure to attempt to
%encapsulate the article, complementing the abstract.  This figure illustrates
%the various print and on-screen layouts for which this paper format has
%been optimized: (a) traditional BMVC print format; (b) on-screen
%single-column format, or large-print paper; (c) full-screen two column, or
%2-up printing. }
%\label{fig:teaser}
%\end{figure}

%\begin{table}
%\begin{center}
%\begin{tabular}{|l|c|}
%\hline
%Method & Frobnability \\
%\hline\hline
%Theirs & Frumpy \\
%Yours & Frobbly \\
%Ours & Makes one's heart Frob\\
%\hline
%\end{tabular}
%\end{center}
%\caption{Results.   Ours is better.}
%\end{table}

\bibliography{egbib}
\end{document}

\maketitle

\begin{abstract}
This supplementary material describes the detailed structures of our hybrid network and the discriminator used in adversarial training. Besides, examples of normal and anomalous events occurring in the 4 benchmark datasets are presented. This document also provides additional information related to our experiments including: (1) evaluation curves of all score maps, (2) visualization of learned convolutional filters in the first layer of the encoder, (3) average training score maps supporting the weight estimation for score combination, and (4) sequences of frame-level scores obtained from our method on some clips sampled from the 4 experimental datasets. 
\end{abstract}

%-------------------------------------------------------------------------
\section{Model architecture details}
The operation in each layer is denoted by its abbreviation as follow: \textbf{conv} for convolution, \textbf{bnorm} for batch-normalization~\cite{Ioffe2015Batch}, \textbf{lrelu} for leaky rectified linear unit~\cite{Maas2013Rectifier}, \textbf{deconv} for deconvolution, and \textbf{fc} for fully-connected layer (i.e. dense layer). This content should be read together with the corresponding section in the main paper for a clear understanding.

\subsection{Hybrid network}
\begin{table}[H]
\vspace{-8pt}
\centering
\scriptsize
\begin{tabular}{ll}
\begin{tabular}[t]{llll}
\hline
\textbf{Component} & \textbf{Layer} & \textbf{Parameter} & \textbf{Output size} \\\hline\hline
%\multirow{1}{*}{Enc/input} & - & - & $10\times10\times3$ \\\hline
\multirow{3}{*}{Enc/block1} & conv & $3\times3~/~1$ & $10\times10\times32$ \\
                            & conv & $3\times3~/~1$ & $10\times10\times64$ \\
                            & lrelu & 0.2 & $10\times10\times64$ \\\hline
\multirow{4}{*}{Enc/block2} & conv & $3\times3~/~2$ & $5\times5\times64$ \\
                            & conv & $3\times3~/~1$ & $5\times5\times128$ \\
                            & bnorm & - & $5\times5\times128$ \\
                            & lrelu & 0.2 & $5\times5\times128$ \\\hline
\multirow{4}{*}{Enc/block3} & conv & $3\times3~/~2$ & $3\times3\times128$ \\
                            & conv & $3\times3~/~1$ & $3\times3\times256$ \\
                            & bnorm & - & $3\times3\times256$ \\
                            & lrelu & 0.2 & $3\times3\times256$ \\\hline
\end{tabular}&
\begin{tabular}[t]{llll}
\hline
\textbf{Component} & \textbf{Layer} & \textbf{Parameter} & \textbf{Output size} \\\hline\hline
\multirow{3}{*}{Dec/block1} & conv & $3\times3~/~1$ & $3\times3\times128$ \\
                            & bnorm & - & $3\times3\times128$ \\
                            & relu & - & $3\times3\times128$ \\\hline
\multirow{5}{*}{Dec/block2} & deconv & $3\times3~/~2$ & $5\times5\times128$ \\
                            & conv & $3\times3~/~1$ & $5\times5\times64$ \\
                            & bnorm & - & $5\times5\times64$ \\
														& dropout & 0.5 & $5\times5\times64$ \\
                            & relu & - & $5\times5\times64$ \\\hline
\multirow{5}{*}{Dec/block3} & deconv & $3\times3~/~2$ & $10\times10\times64$ \\
                            & conv & $3\times3~/~1$ & $10\times10\times32$ \\
                            & bnorm & - & $10\times10\times32$ \\
														& dropout & 0.5 & $10\times10\times32$ \\
                            & relu & - & $10\times10\times32$ \\\hline
\multirow{1}{*}{Dec/output}	& conv & $3\times3~/~1$ & $10\times10\times3$ \\\hline
\end{tabular}
\end{tabular}\vspace{2mm}
\caption{Architecture of the encoder (left) and decoder (right) in our AE (see the main paper for its visualization).}
\label{table:ae}
\end{table}
Our hybrid network can be split into 3 principal parts: encoder, decoder and classification sub-network. The structures of the first two components are presented in Table~\ref{table:ae}. The operations are performed top-down. The architecture of the classification network is given in Table~\ref{table:clf} with some layer inputs from the encoder in Table~\ref{table:ae}. The \textit{Parameter} column indicates the values assigned for the corresponding operation: filter size / stride for convolution, slope for leaky ReLU, and the dropout probability.

% Please add the following required packages to your document preamble:
% \usepackage{multirow}
\begin{table}[H]
\vspace{-8pt}
\centering
\footnotesize
\begin{tabular}{lllll}
\hline
\textbf{Component} &\textbf{Input} & \textbf{Layer} & \textbf{Parameter} & \textbf{Output size} \\\hline\hline
\multirow{2}{*}{Clf/branch1} & \multirow{2}{*}{Enc/block1} & conv & $1\times1~/~1$ & $10\times10\times32$ \\
  & & flatten & - & 3200 \\\hline
\multirow{2}{*}{Clf/branch2} & \multirow{2}{*}{Enc/block2} & conv & $1\times1~/~1$ & $5\times5\times64$ \\
  & & flatten & - & 1600 \\\hline
Clf/branch3 & Enc/block3 & flatten & - & 2304 \\\hline
Clf/feature & \begin{tabular}[c]{@{}l@{}}Clf/branch1\\ Clf/branch2\\ Clf/branch3\end{tabular} & concatenation & - & 7104 \\\hline
\multirow{3}{*}{\begin{tabular}[c]{@{}l@{}}Clf/hidden1\_x\\ Clf/hidden1\_y\end{tabular}} & \multirow{3}{*}{Clf/feature} & fc & - & 1024 \\
 & & relu & - & 1024 \\
 & & dropout & 0.5 & 1024 \\\hline
\multirow{3}{*}{\begin{tabular}[c]{@{}l@{}}Clf/hidden2\_x\\ Clf/hidden2\_y\end{tabular}} & \multirow{3}{*}{\begin{tabular}[c]{@{}l@{}}Clf/hidden1\_x\\ Clf/hidden1\_y\end{tabular}} & fc & - & 1024 \\
 & & relu & - & 1024 \\
 & & dropout & 0.5 & 1024 \\\hline
\begin{tabular}[c]{@{}l@{}}Clf/output\_x\\ Clf/output\_y\end{tabular} & \begin{tabular}[c]{@{}l@{}}Clf/hidden2\_x\\ Clf/hidden2\_y\end{tabular} & softmax & - & \begin{tabular}[c]{@{}l@{}}16\\ 12\end{tabular}\\\hline
\end{tabular}\vspace{2mm}
\caption{Architecture of the classification sub-network.}% Last components are grouped because of their similar structures.}
\label{table:clf}
\end{table}

\subsection{Discriminator}
The structure of our discriminator used during the training stage is presented in Table~\ref{table:disc}. The input of this network is either training samples or reconstructed cuboids obtained from the auto-encoder in the hybrid network. The fully-connected layer in the last block corresponds to a weighted sum of its 4 input units.
\begin{table}[H]
\vspace{-8pt}
\centering
\footnotesize
\begin{tabular}[t]{llll}
\hline
\textbf{Component} & \textbf{Layer} & \textbf{Parameter} & \textbf{Output size} \\\hline\hline
\multirow{1}{*}{Disc/input} & - & - & $10\times10\times3$ \\\hline
\multirow{2}{*}{Disc/block1} & conv & $3\times3~/~2$ & $5\times5\times32$ \\
                            & lrelu & 0.2 & $5\times5\times32$ \\\hline
\multirow{3}{*}{Disc/block2} & conv & $3\times3~/~2$ & $3\times3\times64$ \\
                            & bnorm & - & $3\times3\times64$ \\
                            & lrelu & 0.2 & $3\times3\times64$ \\\hline
\multirow{3}{*}{Disc/block3} & conv & $3\times3~/~2$ & $2\times2\times128$ \\
                            & bnorm & - & $2\times2\times128$ \\
                            & lrelu & 0.2 & $2\times2\times128$ \\\hline
\multirow{4}{*}{Disc/block4} & conv & $1\times1~/~1$ & $2\times2\times1$ \\
                            & bnorm & - & $2\times2\times1$ \\
                            & fc & - & 1 \\
														& sigmoid & - & 1 \\\hline
\end{tabular}\vspace{2mm}
\caption{Architecture of the discriminator (see the main paper for its visualization).}
\label{table:disc}
\end{table}

\section{Normal and abnormal events in benchmark datasets}
\begin{figure}[H]
\centering
\includegraphics[width=0.34\textwidth]{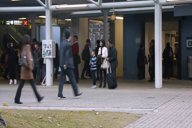}\hspace{3.3mm}
\includegraphics[width=0.34\textwidth]{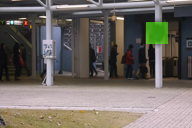}\\
(a) Avenue dataset (anomaly: a bag is falling).\\\vspace{3.2mm}
\includegraphics[width=0.34\textwidth]{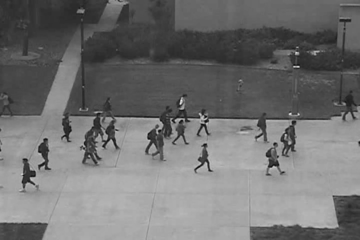}\hspace{3.3mm}
\includegraphics[width=0.34\textwidth]{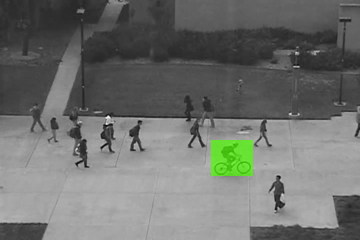}\\
(b) Ped2 dataset (anomaly: bicycle running).\\\vspace{3.2mm}
\includegraphics[width=0.34\textwidth]{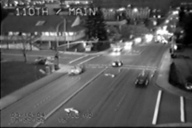}\hspace{3.3mm}
\includegraphics[width=0.34\textwidth]{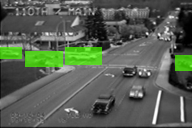}\\
(c) Belleview dataset (anomaly: cars running on side roads).\\\vspace{3.2mm}
\includegraphics[width=0.34\textwidth]{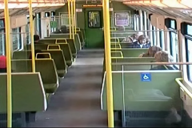}\hspace{3.3mm}
\includegraphics[width=0.34\textwidth]{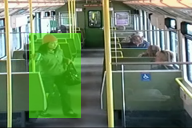}\\
(d) Traffic-Train dataset (anomaly: passenger movement).\\\vspace{3.2mm}
\includegraphics[width=0.34\textwidth]{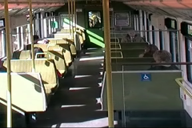}\hspace{3.3mm}
\includegraphics[width=0.34\textwidth]{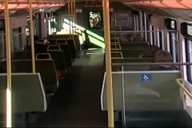}\\
(e) The lighting condition drastically changed in the Traffic-Train dataset.
\caption{(a)-(d) Normal (left) and anomalous (right) events in the 4 benchmark datasets where the image regions of anomaly are highlighted by green boxes; (e) Challenge of lighting change in the Traffic-Train dataset. Best viewed in color.}
\label{fig:example}
\end{figure}

Figure~\ref{fig:example} presents video frames sampled from some normal and abnormal events (annotated in the provided ground truths) in the Avenue, Ped2, Belleview and Traffic-Train datasets. An example of different lighting conditions in the last dataset is also shown in this figure.

\section{Evaluation curves of all score maps}
\begin{figure}[H]
\centering
\includegraphics[width=0.42\textwidth]{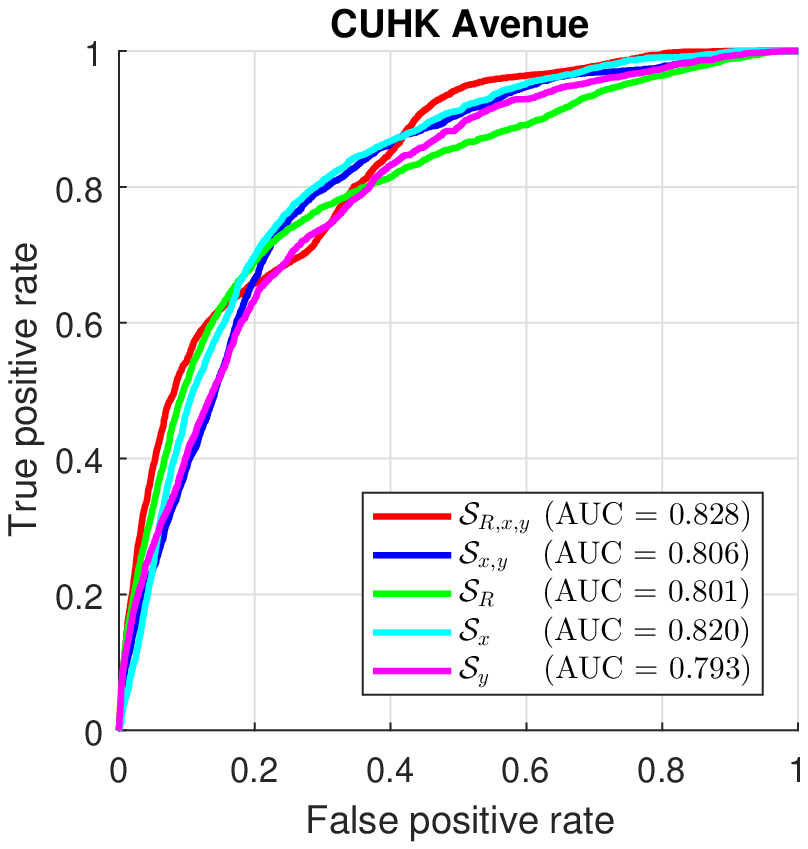}\hspace{5mm}
\includegraphics[width=0.42\textwidth]{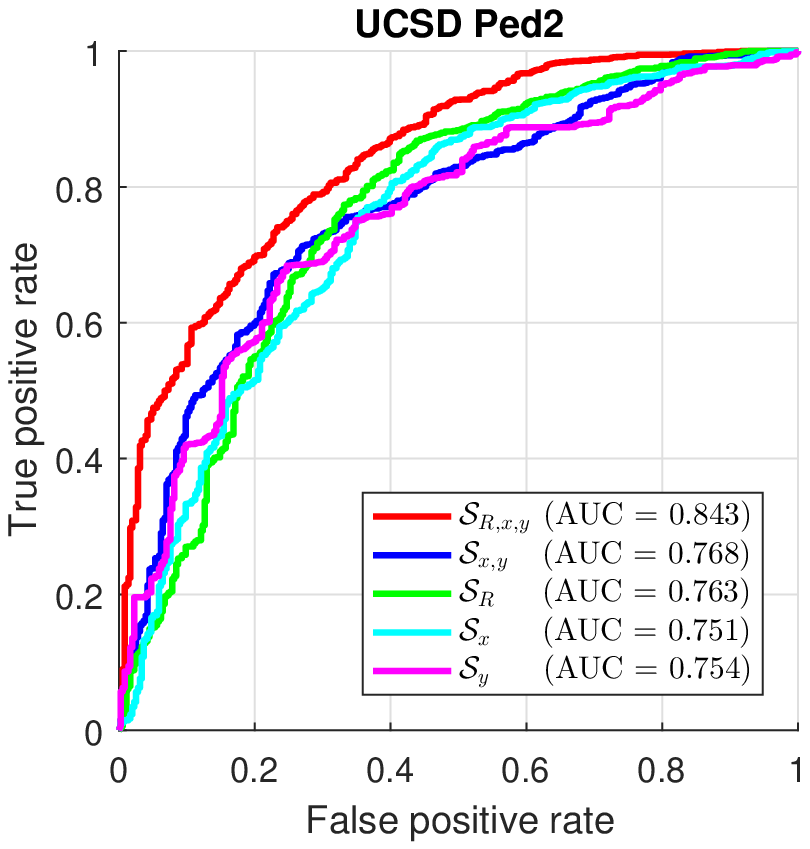}\\\vspace{1.7mm}
\includegraphics[width=0.42\textwidth]{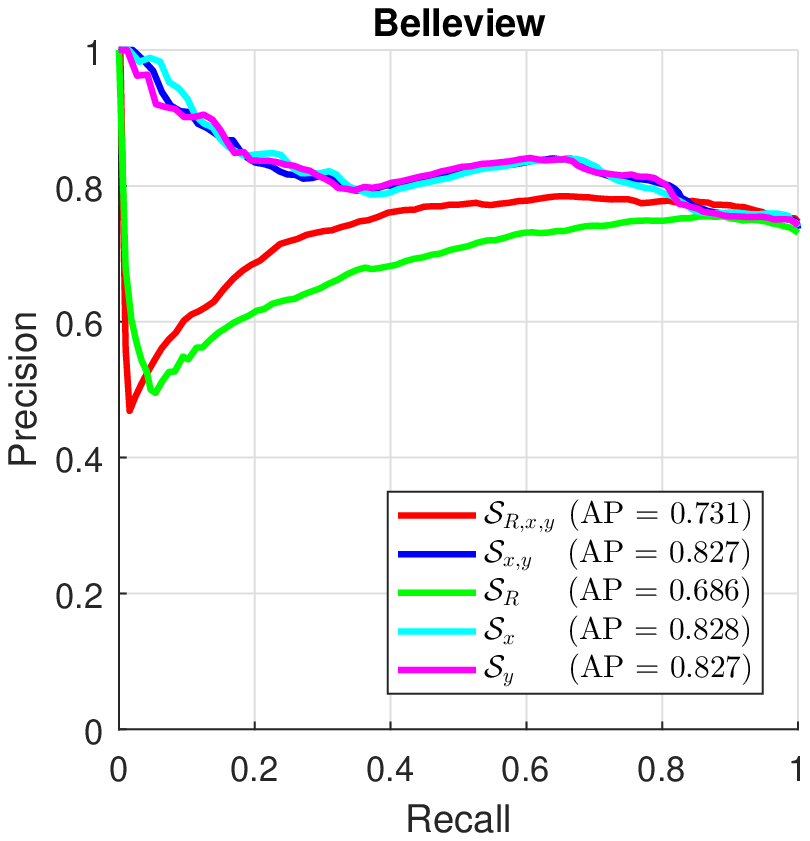}\hspace{5mm}
\includegraphics[width=0.42\textwidth]{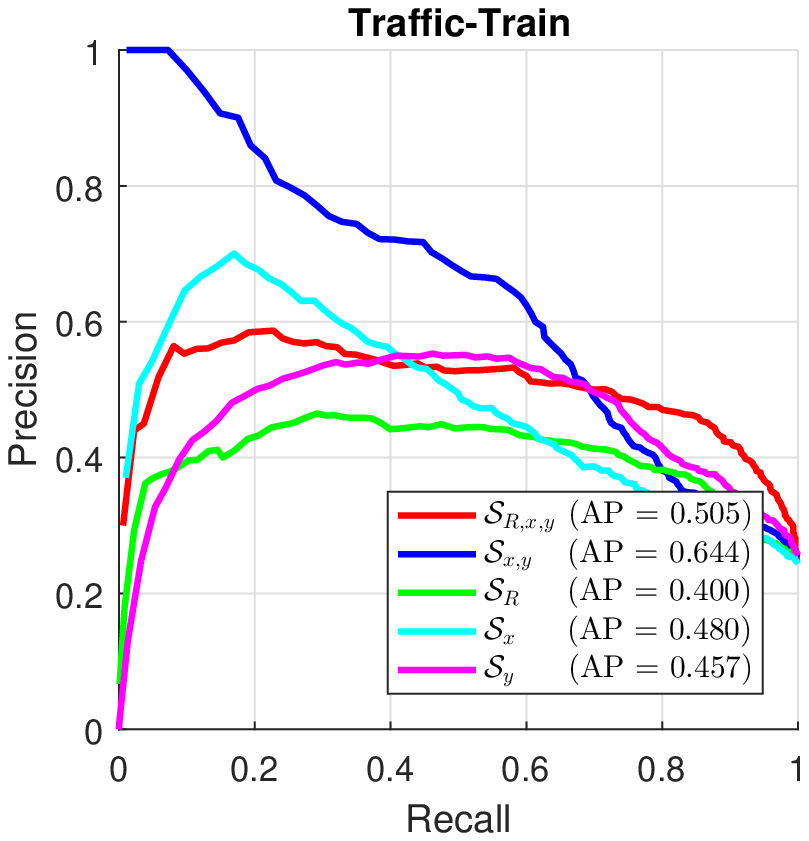}
\caption{Evaluation curves provided by individual score maps and their combinations in our experiments. ROC curves are used for Avenue and Ped2 datasets, and PR curves are for the others. Best viewed in color.}
\label{fig:intra_curves}
\end{figure}

Figure~\ref{fig:intra_curves} shows the evaluation curves corresponding to the 5 score maps introduced in our work. These results were obtained with the decoder in the hybrid network. For the Avenue and Ped2 datasets, the score map $\mathcal{S}_R$ given by the reconstruction provided competitive results compared with the classification sub-network. The combination of the three partial maps $\mathcal{S}_R$, $\mathcal{S}_x$ and $\mathcal{S}_y$ hence improved the anomaly measurement.

On the contrary, the low efficiency of $\mathcal{S}_R$ in the Belleview and Traffic-Train datasets led to a low performance of $\mathcal{S}_{R,x,y}$. The removal of $\mathcal{S}_R$ in the combination was thus an appropriate choice since $\mathcal{S}_{x,y}$ provided much better evaluation results.

\section{Learned convolutional filters}
\begin{figure}[t]
\centering
~~\footnotesize{Learned with decoder}~~~~~~~~~~~~~~~~~~~~~~~~\footnotesize{Learned without decoder}~~~~~~~~~~~~~~~~~~~~~~~~~~~~~~~~~~~\footnotesize{Difference}~~~~~~~~~~~\\\vspace{1.7mm}
\includegraphics[width=\textwidth]{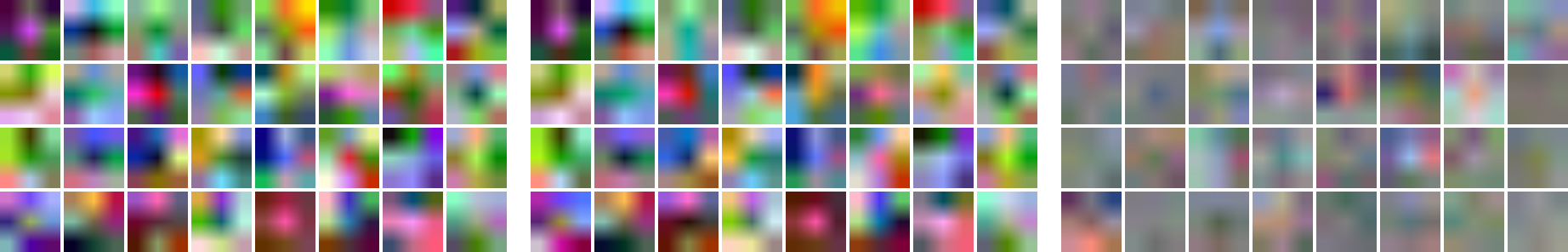}\\(a) CUHK Avenue dataset\\\vspace{1.7mm}
\includegraphics[width=\textwidth]{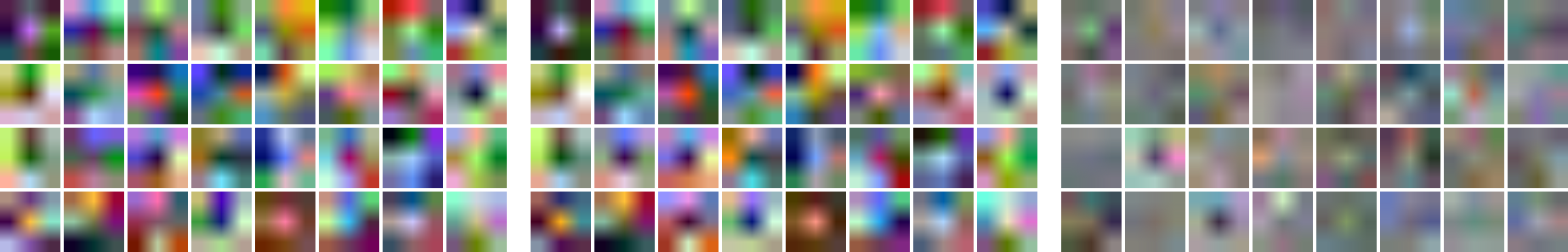}\\(b) UCSD Ped2 dataset\\\vspace{1.7mm}
\includegraphics[width=\textwidth]{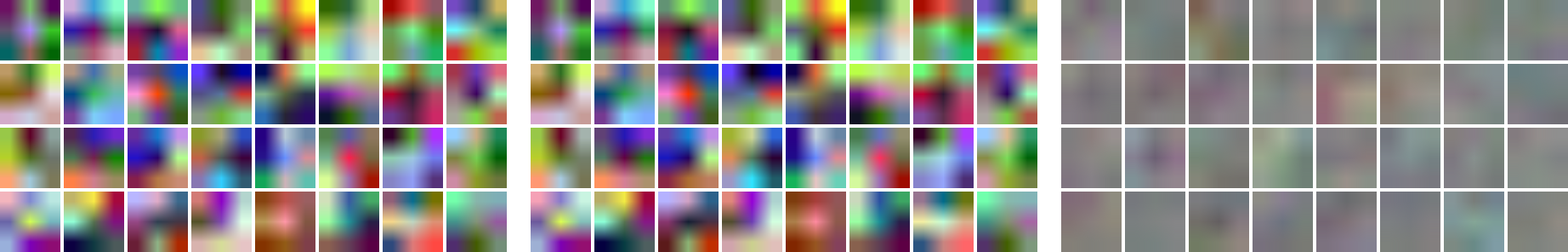}\\(c) Belleview dataset\\\vspace{1.7mm}
\includegraphics[width=\textwidth]{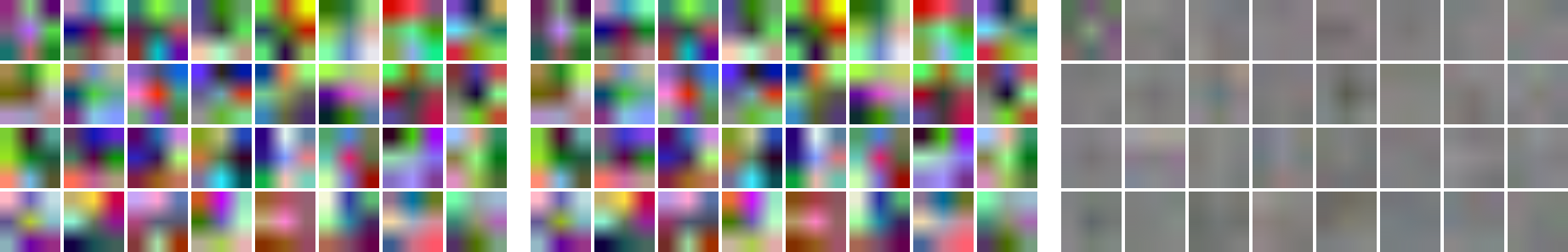}\\(d) Traffic-Train dataset
\caption{Learned filters in the first convolutional layer obtained from our experiments on the 4 benchmark datasets. Best viewed in color.}
\label{fig:filters}
\end{figure}

In order to provide a visual comparison between the hybrid network and its modification without decoder, we present the learned filters in the first convolutional layer (i.e. 32 filters in the component Enc/block1 in Table~\ref{table:ae}). Specifically, they show the low-level characteristics that the models expect for an individual cuboid. The learned filters are respectively visualized in Fig.~\ref{fig:filters} for the Avenue, Ped2, Belleview and Traffic-Train datasets.

The original size of each filter is $3\times3\times3$, in which the last value indicates the number of concatenated frames used for cuboid formation. They are up-scaled in the figures to provide a better visual understanding of weight distribution. Each collection of filters corresponding to a dataset is represented as 3 columns. The first and second columns are the filters corresponding to the hybrid network with and without the decoder, and the third one visualizes their difference. The 32 filters are arranged in the same order for all datasets.

The first two columns in Fig.~\ref{fig:filters} show that the overall structures of learned filters in the first convolutional layer are similar for the two models (with/without decoder). The third column shows the difference which is less significant for Belleview and Traffic-Train. This means that the effect (i.e. contribution) of the decoder in the hybrid network would be reduced if its reconstruction task encounters difficulties (due to the low-quality frames in Belleview and the various lighting conditions in Traffic-Train). However, the decoder is still useful for anomaly detection since its use improves the efficiency (see Section 4.4 in the main paper).

\section{Average training score maps for weight estimation}
\begin{figure}[t]
\centering
~~~~~~~~~~~~~\footnotesize{Example}~~~~~~~~~~~~~~~~~~~~~~~~\footnotesize{$\mathcal{S}_R$-based score map}~~~~~~~~~~~~~~~\footnotesize{$\mathcal{S}_x$-based score map}~~~~~~~~~~~~~~~\footnotesize{$\mathcal{S}_y$-based score map}~~~~~~~~\\\vspace{1.7mm}
\includegraphics[width=0.24\textwidth,height=0.18\textwidth]{images/examples/Avenue_normal.png}~
\includegraphics[width=0.24\textwidth]{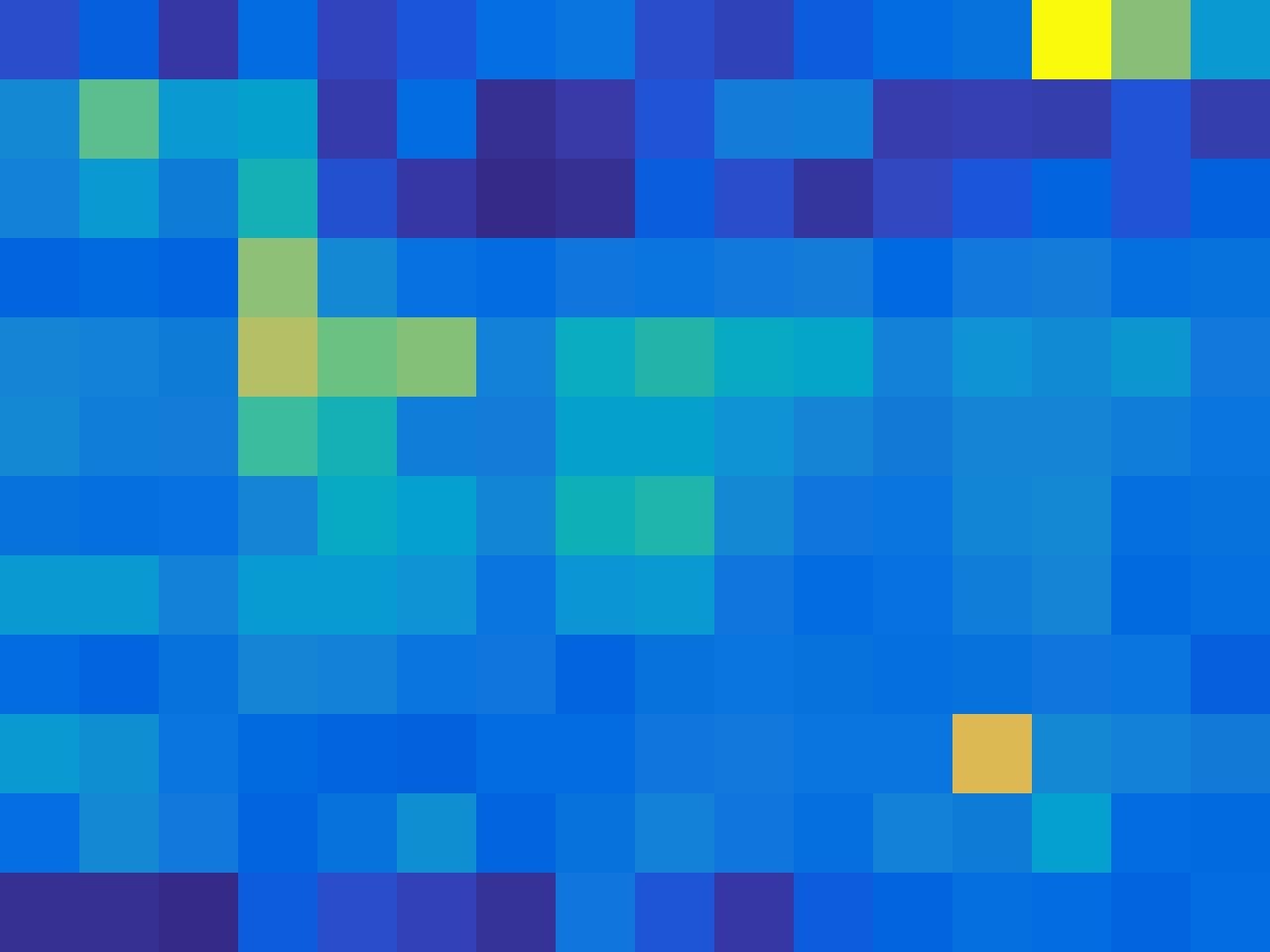}~~\includegraphics[width=0.24\textwidth]{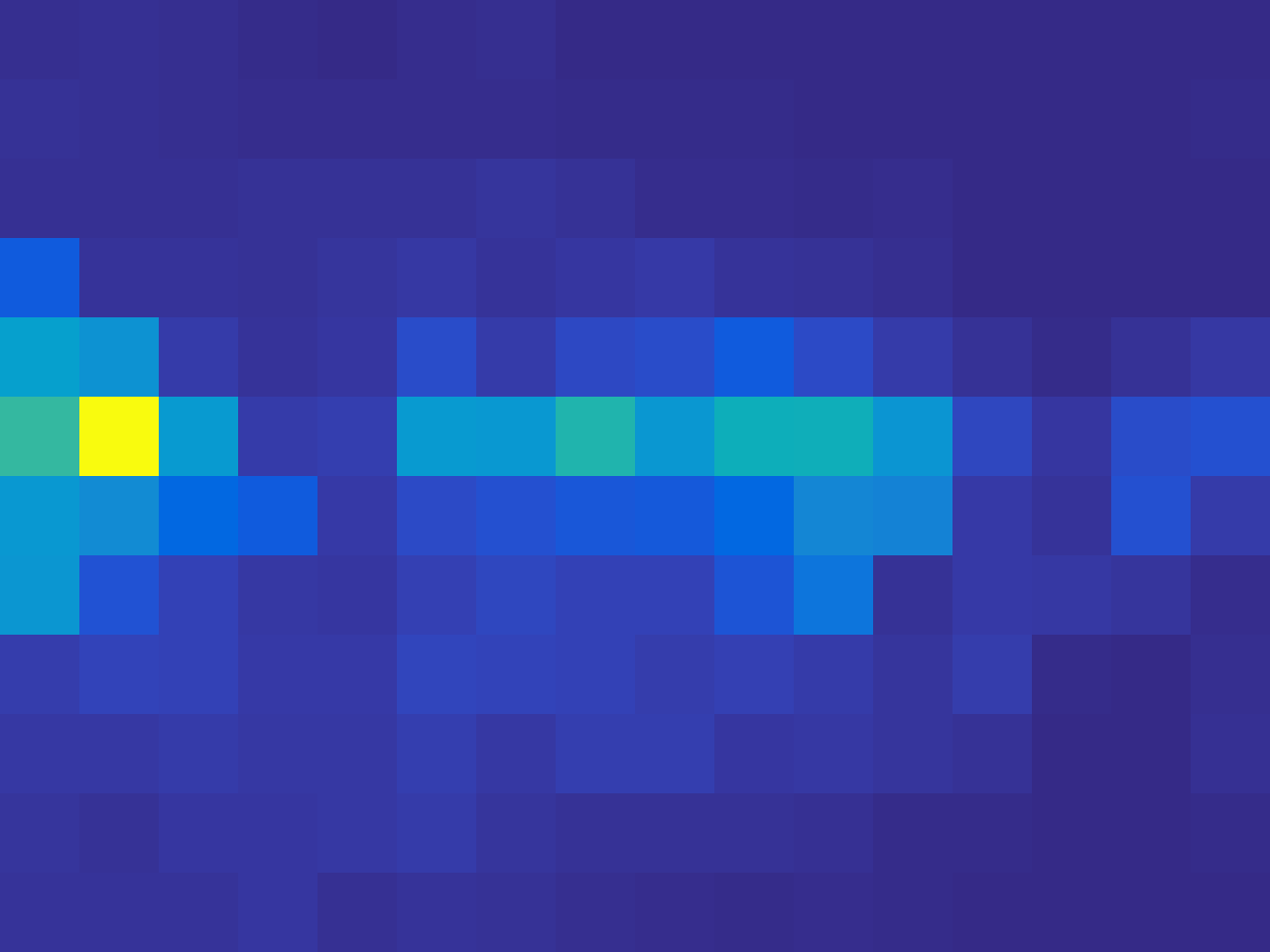}~~\includegraphics[width=0.24\textwidth]{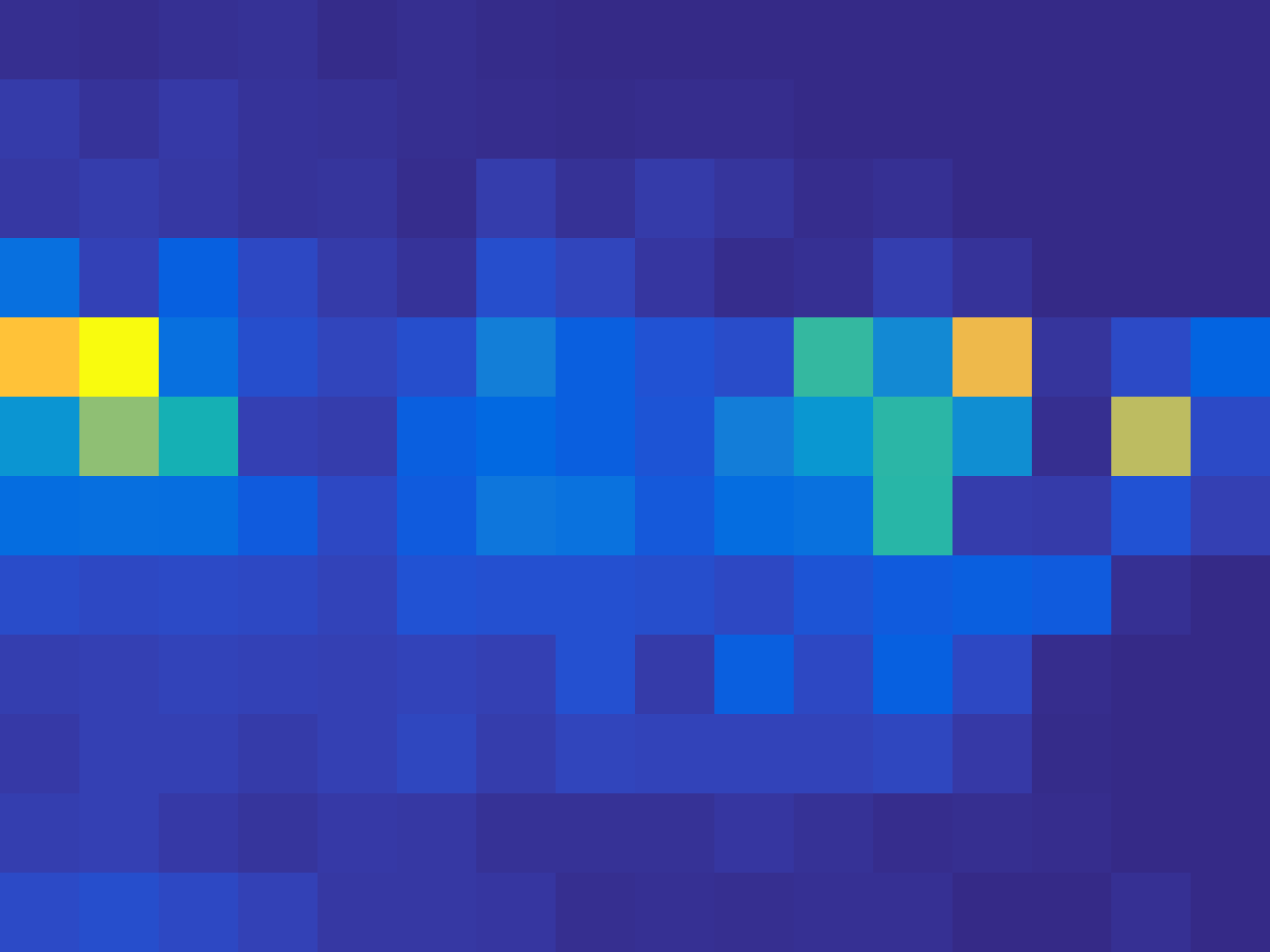}\\(a) CUHK Avenue dataset\\\vspace{1.7mm}
\includegraphics[width=0.24\textwidth,height=0.18\textwidth]{images/examples/Ped2_normal.png}~
\includegraphics[width=0.24\textwidth]{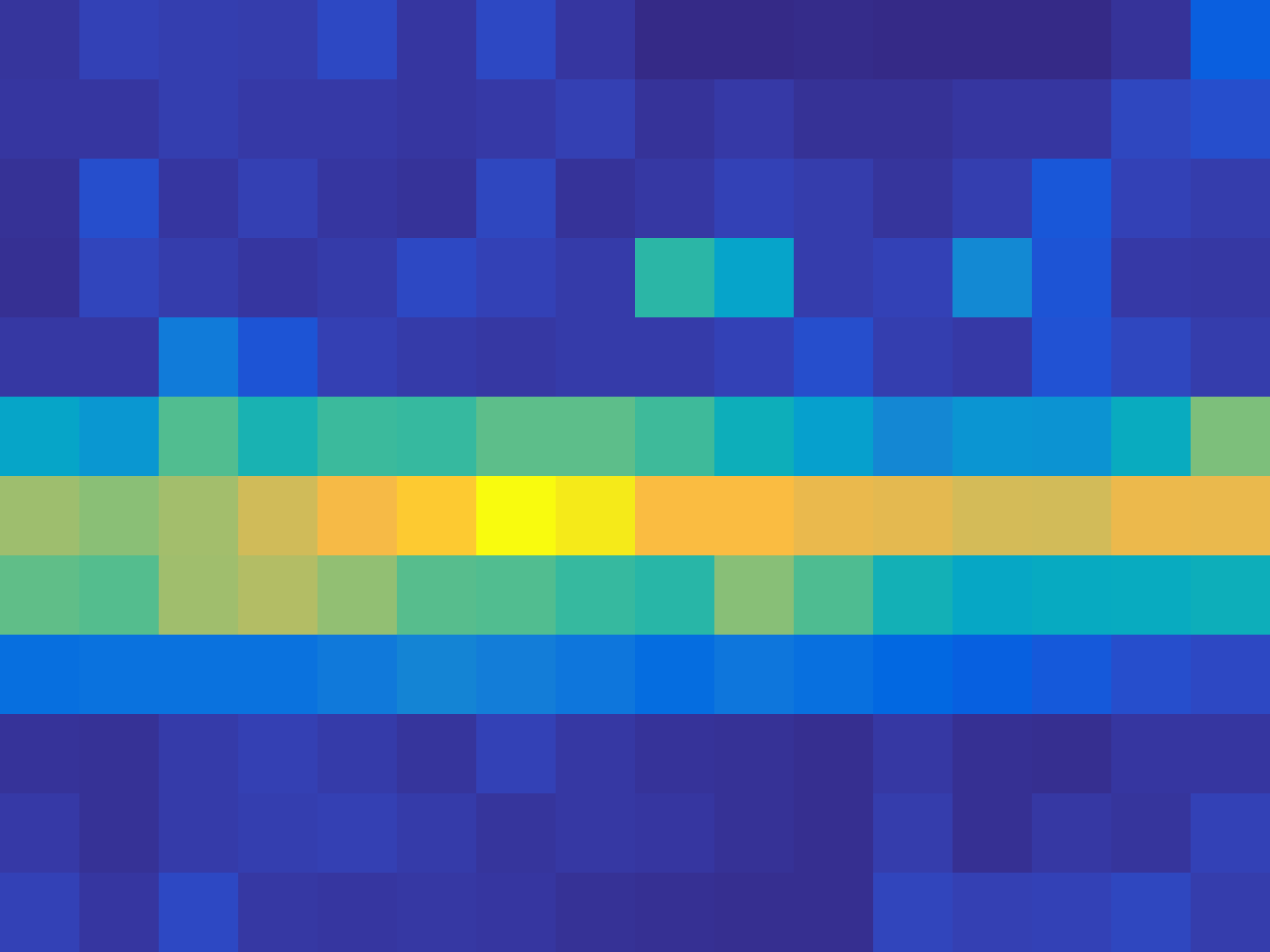}~~\includegraphics[width=0.24\textwidth]{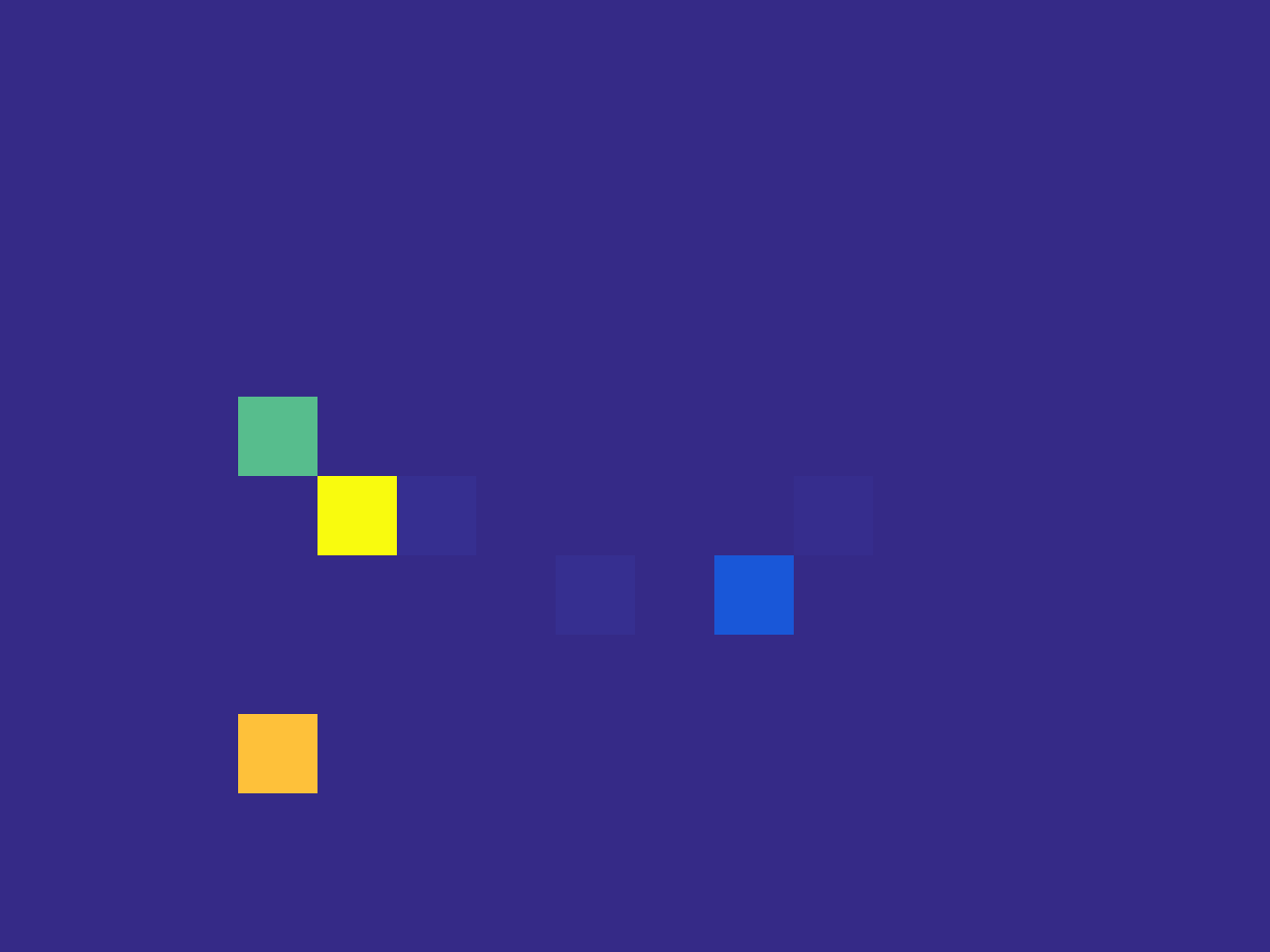}~~\includegraphics[width=0.24\textwidth]{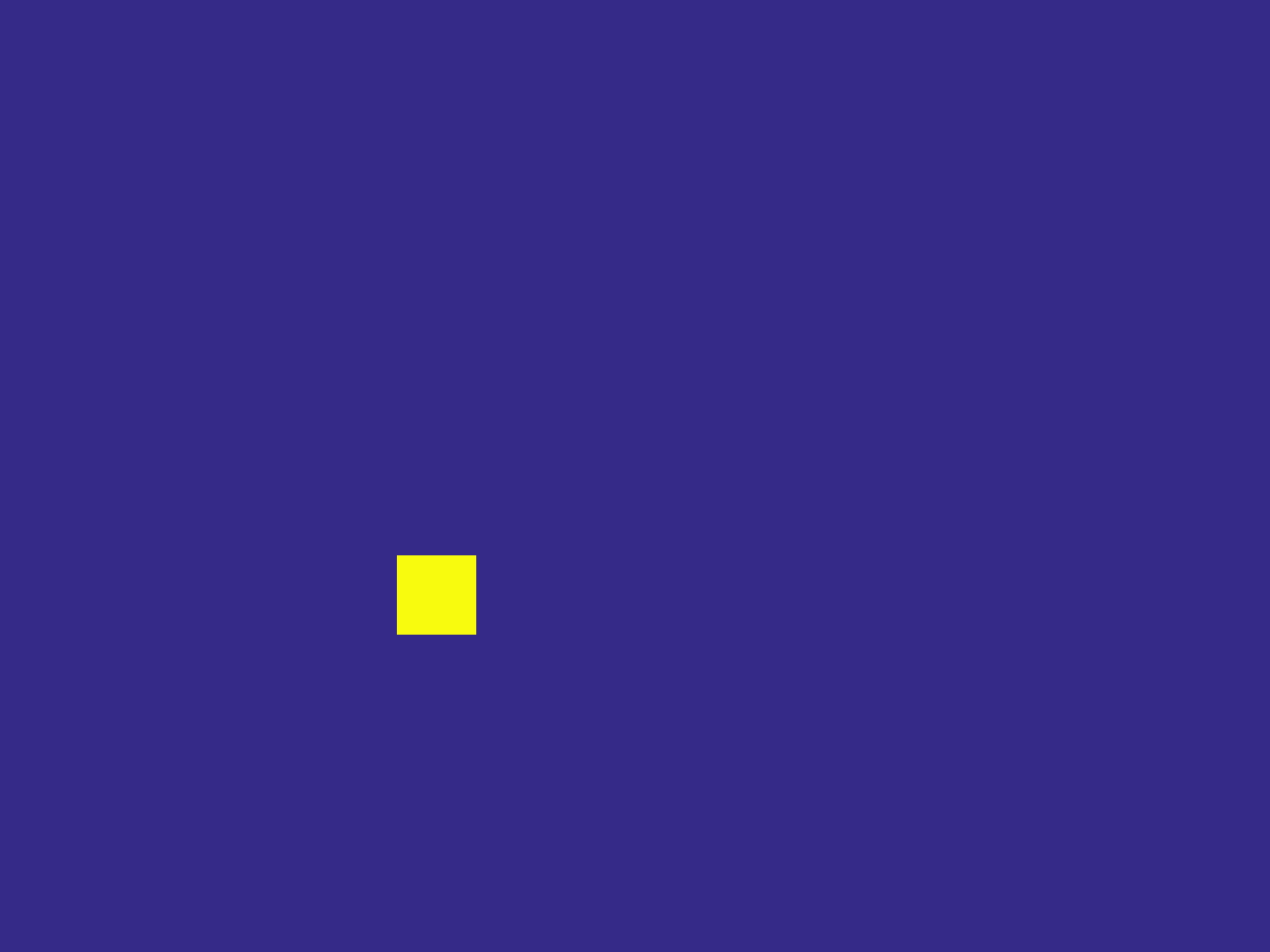}\\(b) UCSD Ped2 dataset\\\vspace{1.7mm}
\includegraphics[width=0.24\textwidth,height=0.18\textwidth]{images/examples/Belleview_normal.png}~
\includegraphics[width=0.24\textwidth]{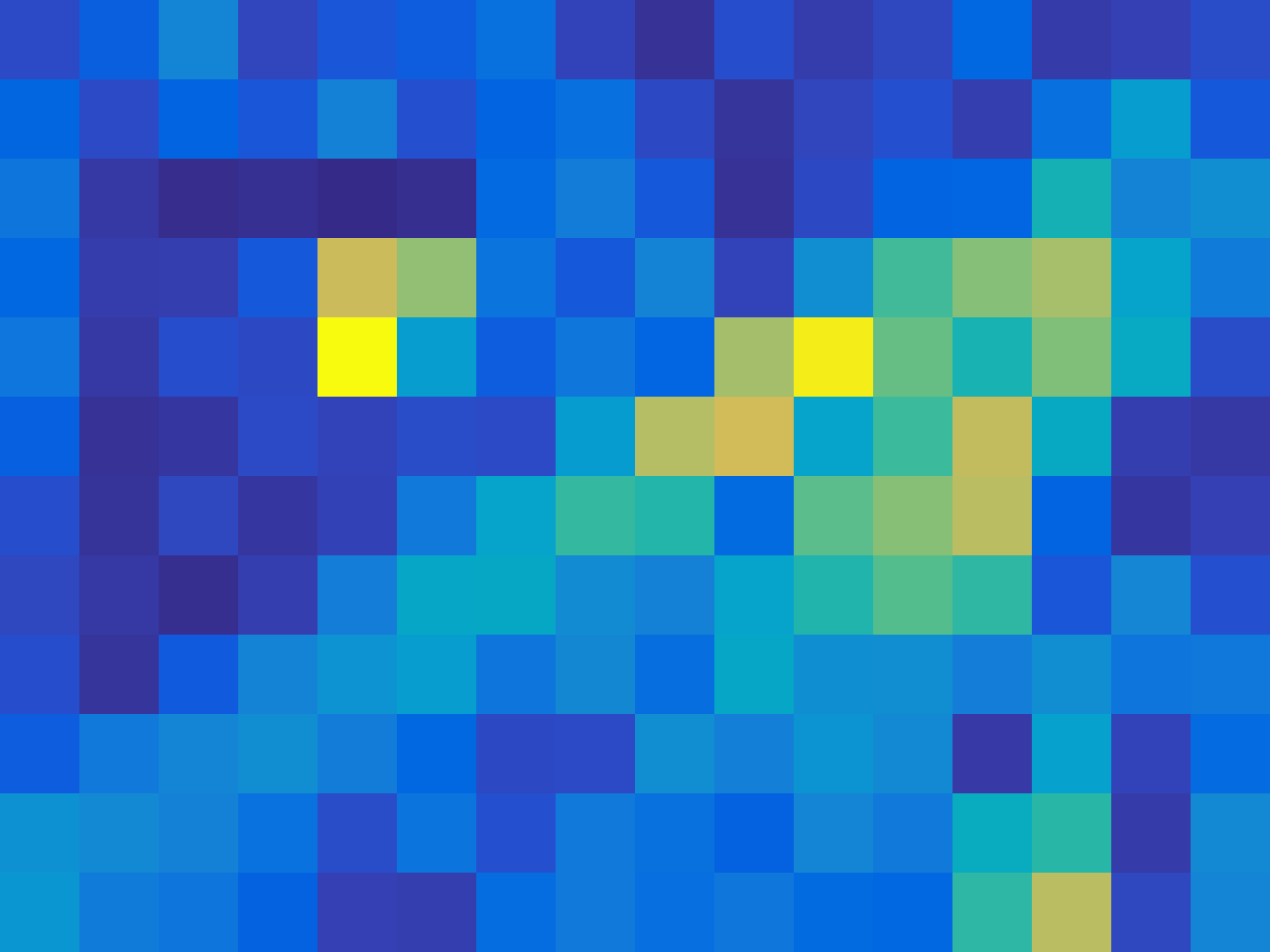}~~\includegraphics[width=0.24\textwidth]{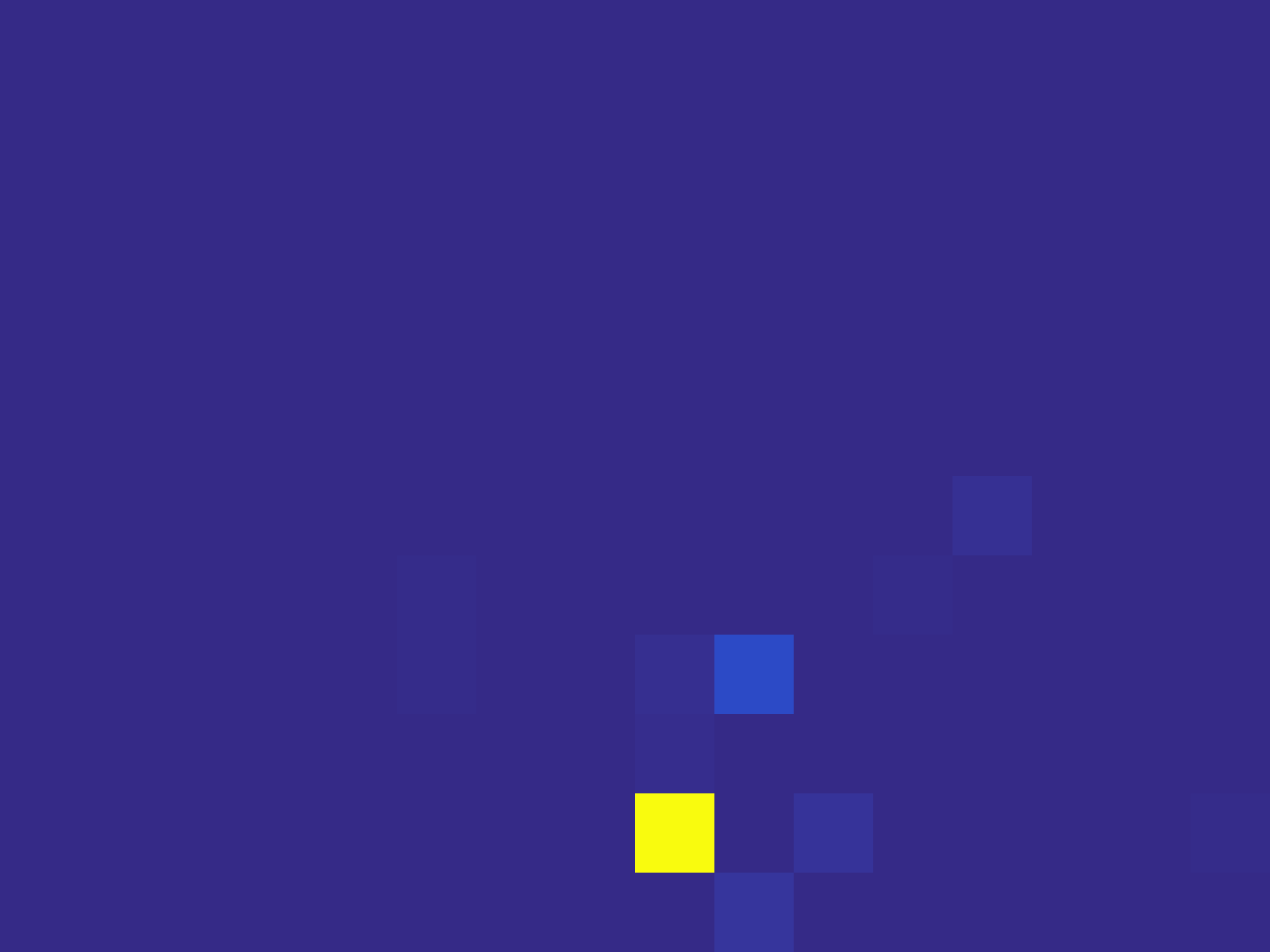}~~\includegraphics[width=0.24\textwidth]{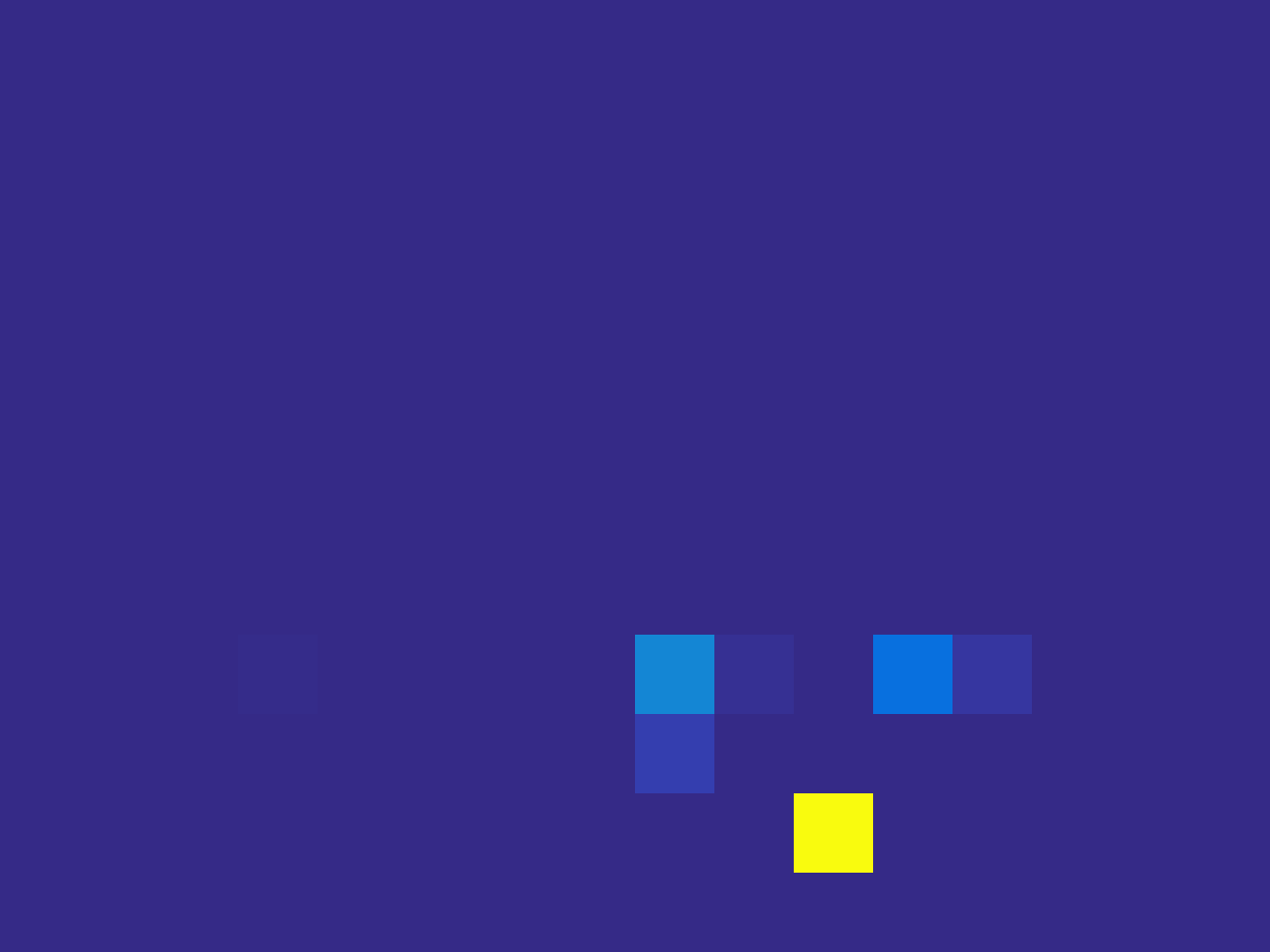}\\(c) Belleview dataset\\\vspace{1.7mm}
\includegraphics[width=0.24\textwidth,height=0.18\textwidth]{images/examples/0342.png}~
\includegraphics[width=0.24\textwidth]{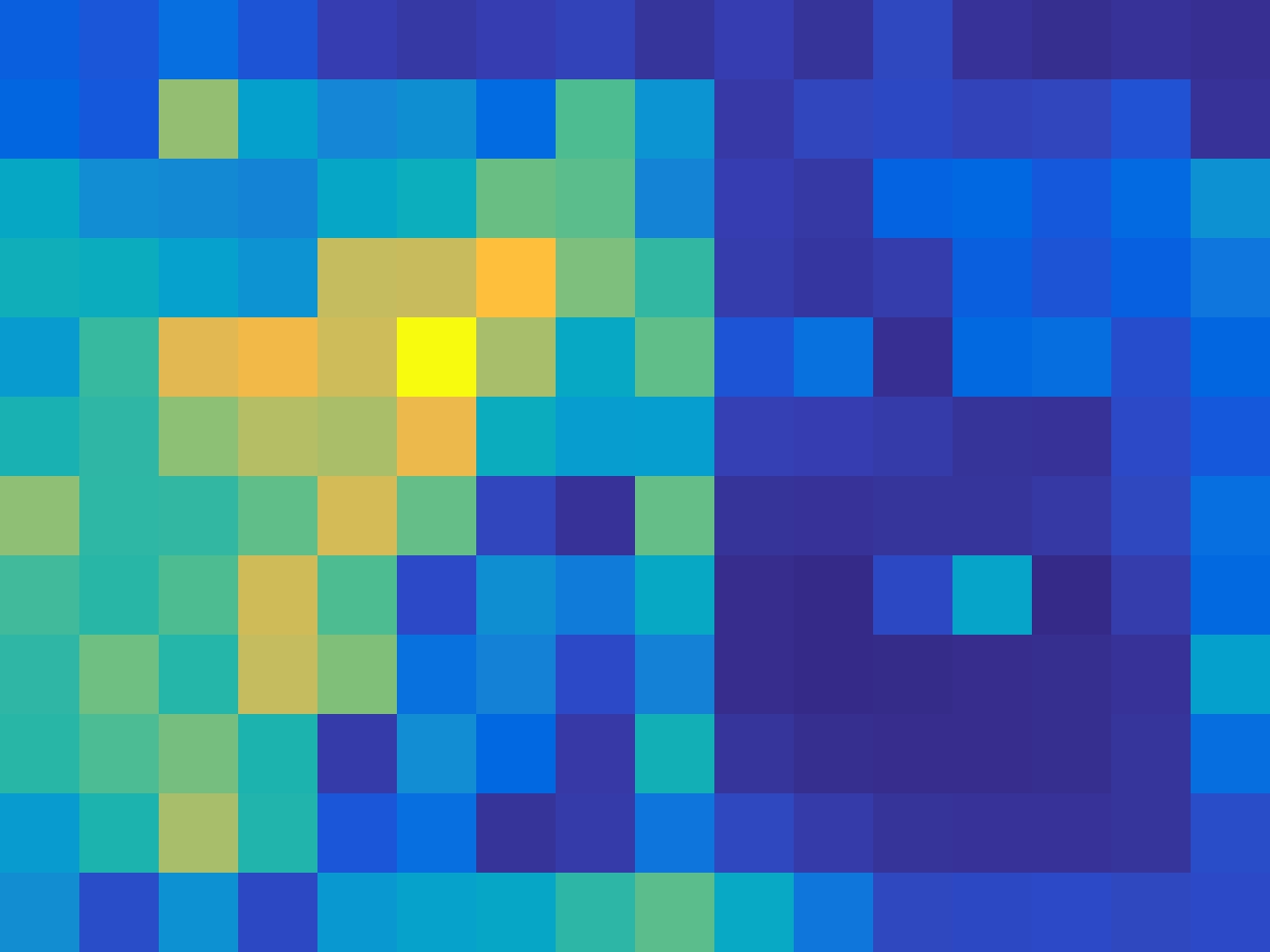}~~\includegraphics[width=0.24\textwidth]{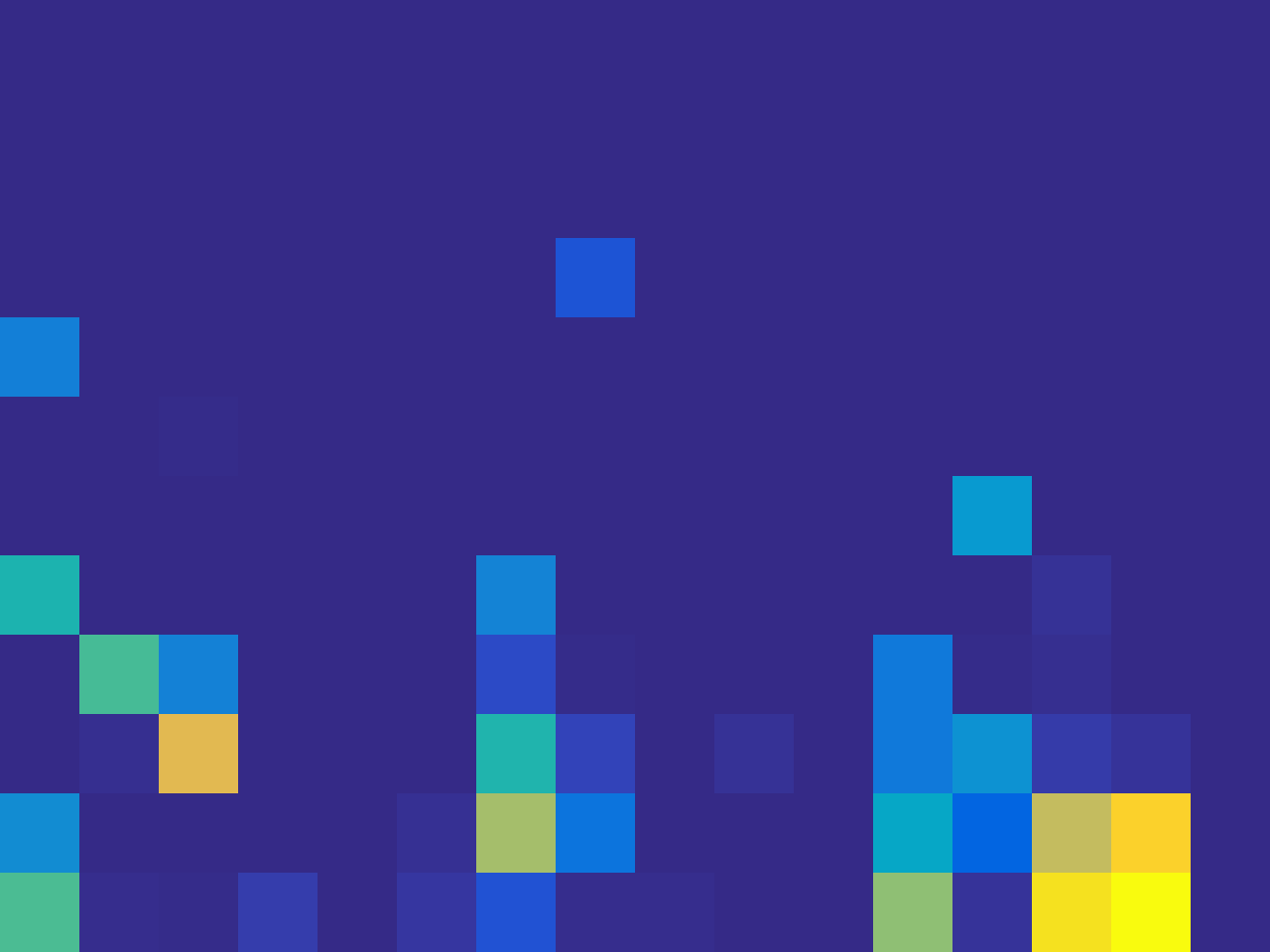}~~\includegraphics[width=0.24\textwidth]{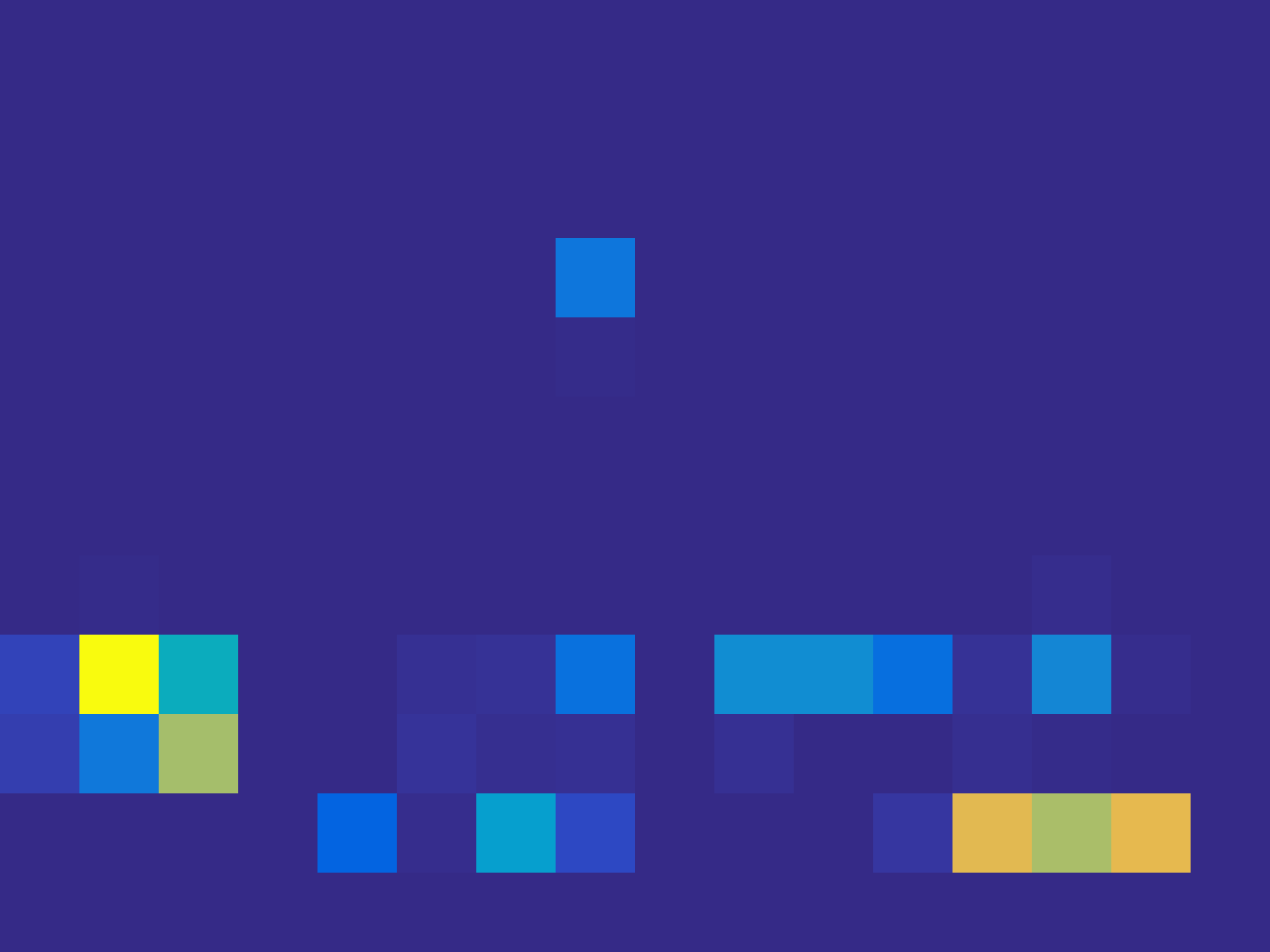}\\(d) Traffic-Train dataset\\\vspace{1.7mm}
color mapping: 0~\includegraphics[width=0.6\textwidth,height=0.01\textwidth]{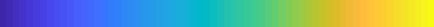}~1
\caption{Average training score maps (for weight estimation) computed in our experiments on the 4 benchmark datasets. Examples of normal frames are presented to provide a visual correspondence between a scene overview and its average training score maps. Best viewed in color.}
\label{fig:weights}
\end{figure}

As presented in the main paper, a combination of individual score maps is performed by a weighted sum, in which the weight is estimated according to training data:
\begin{equation}
	\mathcal{S}_{R,x,y}(\mathbf{c})=\mathlarger{‎‎\mathlarger{‎‎\sum}}_{k\in\{R,x,y\}}\overbrace{\bigg[1-\underbrace{\frac{1}{\|\mathbf{T}\|}‎‎\mathlarger{\sum}_{\mathbf{x}\in\mathbf{T}}\mathcal{S}_k(\mathbf{x})}_\text{average training score}\bigg]}^\text{weight}\mathcal{S}_k(\mathbf{c})
	\label{eq:score_map}
\end{equation}
where $\mathcal{S}_k(\mathbf{c})$ denotes scores estimated on cuboid $\mathbf{c}$, and $\mathbf{T}$ indicates the collection of cuboids at the spatial location $(\mathbf{c}_x,\mathbf{c}_y)$ in the training set consisting of only normal events. The three average training score maps estimated from the entire training data for each experimented dataset are presented in Fig.~\ref{fig:weights}. They are normalized to emphasize the effect of each cuboid location.

For the reconstruction score map $\mathcal{S}_R$, the areas showing movement had higher scores compared with static regions since their reconstructions were more challenging. The score maps $\mathcal{S}_x$ and $\mathcal{S}_y$ provided lower scores for most spatial locations with some sparse higher values, except for the Avenue dataset where higher scores were more numerous and spread out. Higher scores occurred due to the difficulty in classifying very similar cuboids in crowded or uniform regions. Notice that a high value in these average training score maps corresponds to a significant mistake and thus leads to a low weight for the combination in Eq.~(\ref{eq:score_map}).

\section{Examples of score sequences}
\begin{figure}[t]
%\centering
%\includegraphics[width=0.7\textwidth]{images/Avenue_clip_0_Rxy.pdf}
\begin{picture}(450,210)
\put(40,-18){\includegraphics[width=0.79\textwidth]{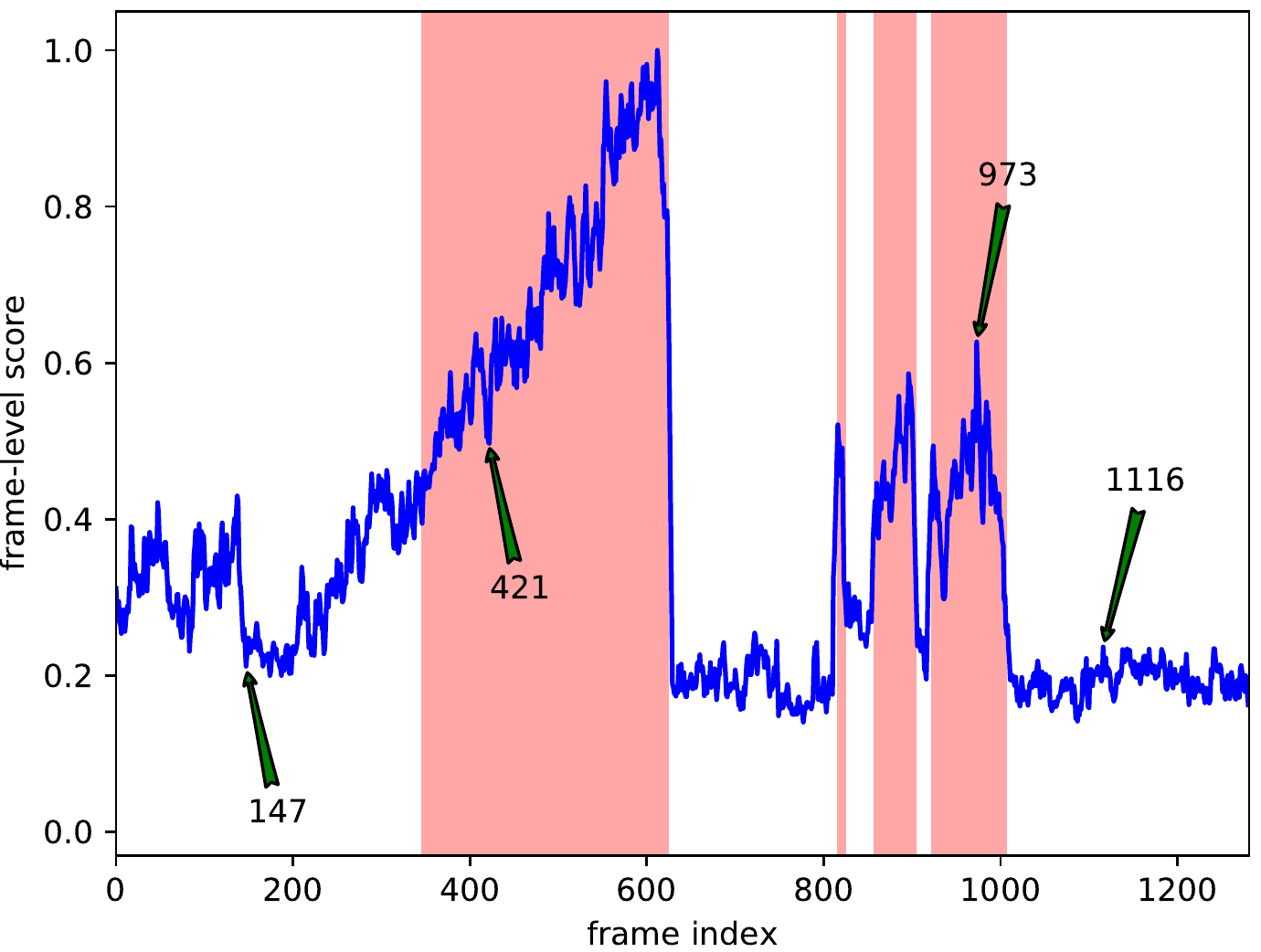}}
\put(72,-18){\includegraphics[scale=0.34]{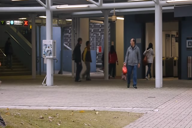}}
\put(126,28){\includegraphics[scale=0.34]{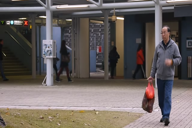}}
\put(237,147){\includegraphics[scale=0.34]{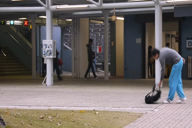}}
\put(275,70){\includegraphics[scale=0.34]{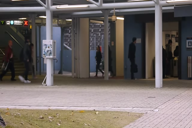}}
\end{picture}
\caption{Frame-scores estimated on Avenue dataset using $\mathcal{S}_{R,x,y}$. Best viewed in color.}
\label{fig:scores_Avenue}
\end{figure}

\begin{figure}[t]
%\centering
%\includegraphics[width=0.72\textwidth]{images/UCSDped2_clip_3_Rxy.pdf}
\begin{picture}(450,210)
\put(37,-18){\includegraphics[width=0.79\textwidth]{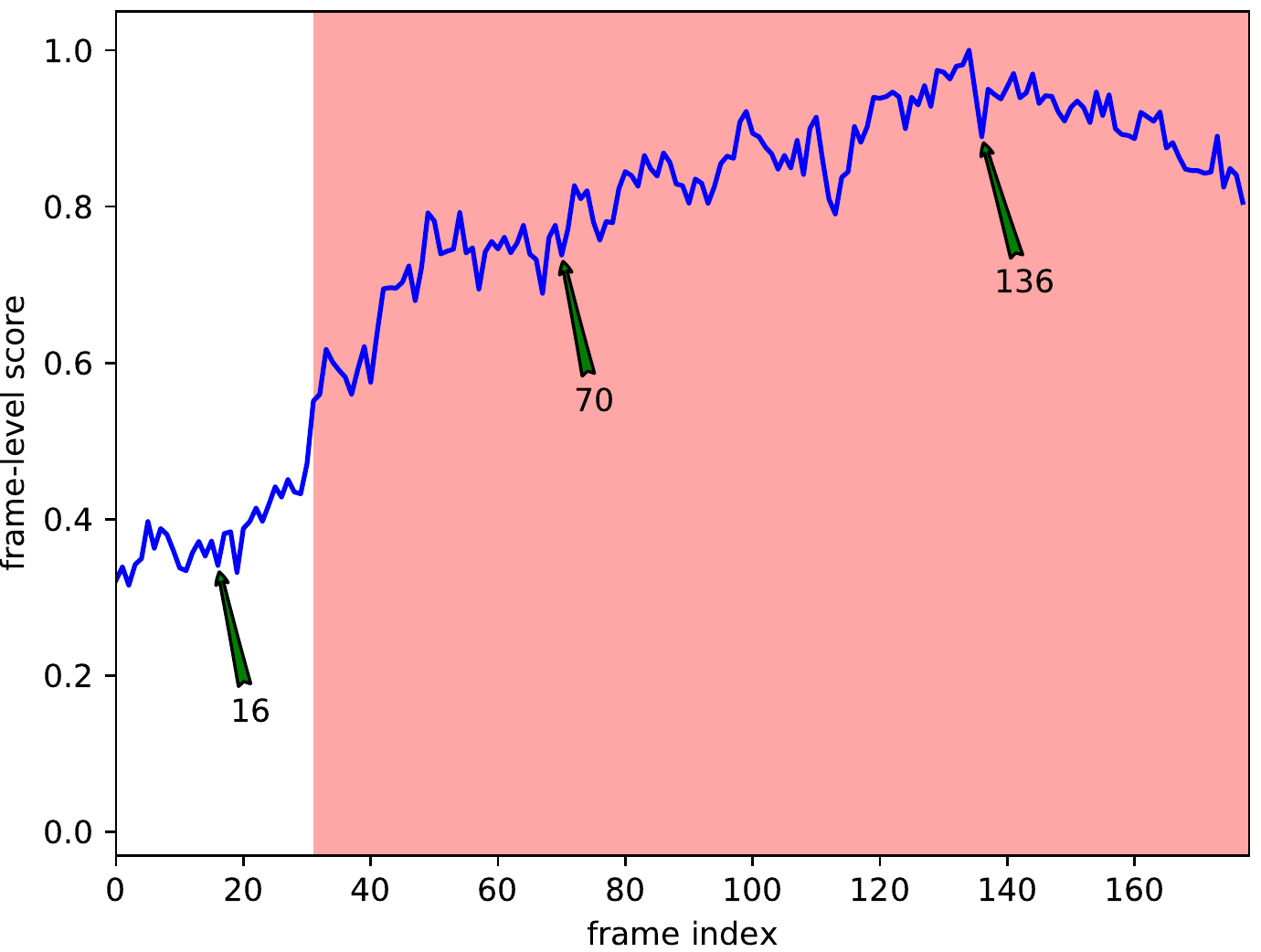}}
\put(75,8){\includegraphics[scale=0.19]{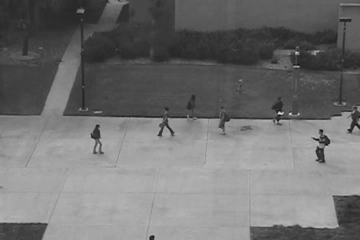}}
\put(150,70){\includegraphics[scale=0.19]{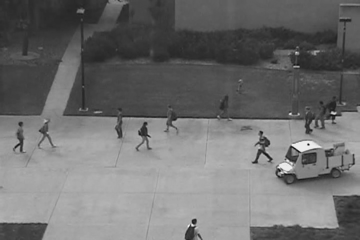}}
\put(240,100){\includegraphics[scale=0.19]{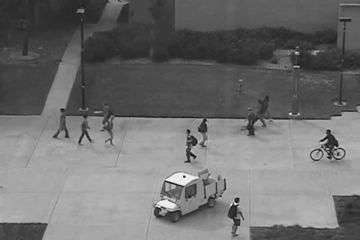}}
\end{picture}
\caption{Frame-scores estimated on Ped2 dataset using $\mathcal{S}_{R,x,y}$. Best viewed in color.}
\label{fig:scores_Ped2}
\end{figure}

\begin{figure}[t]
%\centering
%\includegraphics[width=0.74\textwidth]{images/Belleview_clip_0_xy.pdf}
\begin{picture}(450,210)
\put(43,-18){\includegraphics[width=0.79\textwidth]{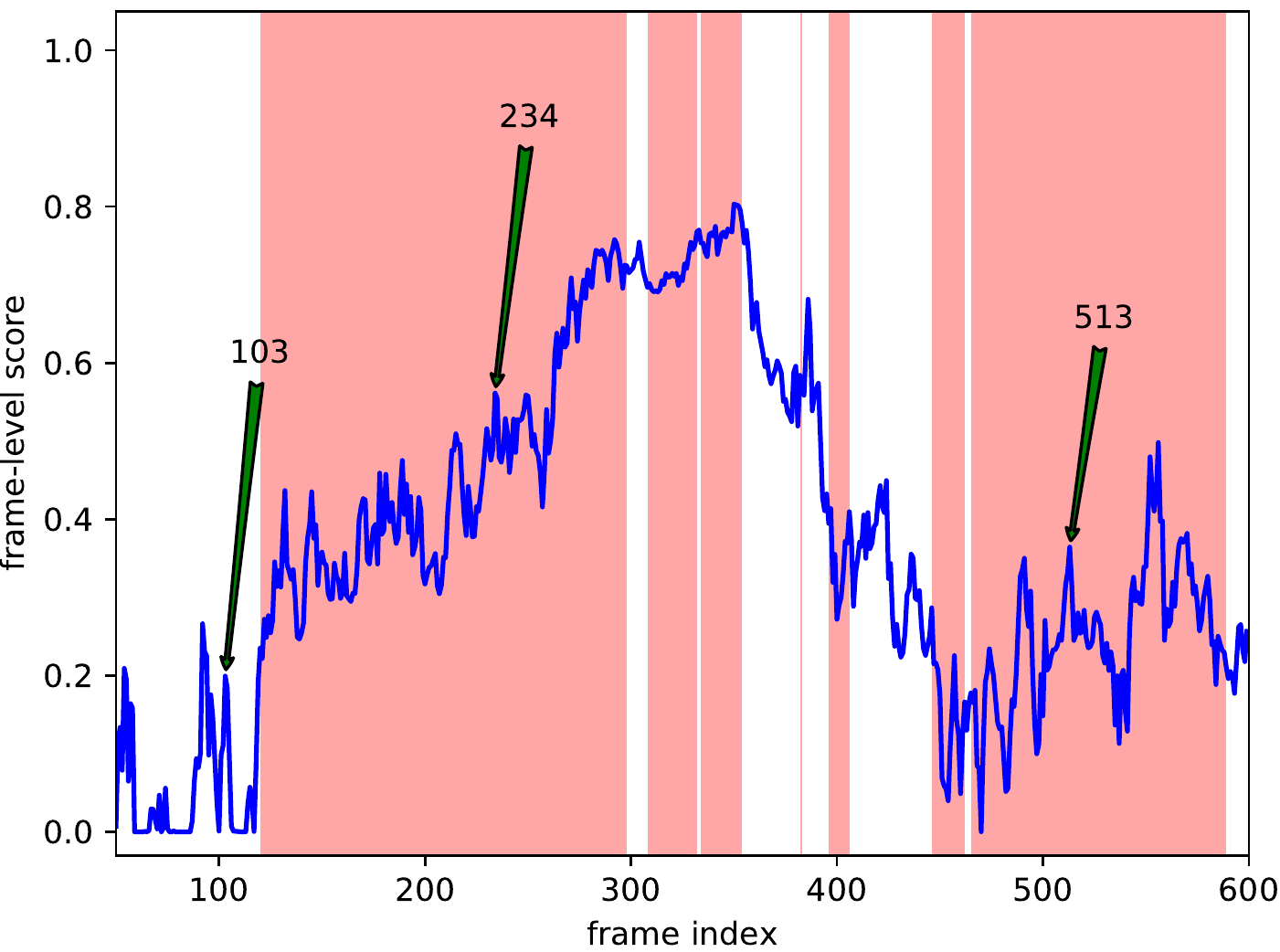}}
\put(72,98){\includegraphics[scale=0.32]{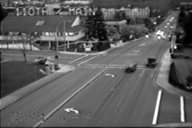}}
\put(135,147){\includegraphics[scale=0.32]{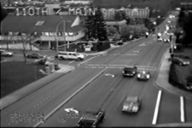}}
\put(262,107){\includegraphics[scale=0.32]{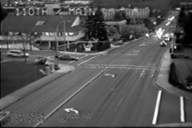}}
\end{picture}
\caption{Frame-scores estimated on Belleview dataset using $\mathcal{S}_{x,y}$. Best viewed in color.}
\label{fig:scores_Belleview}
\end{figure}

\begin{figure}[t]
%\centering
%\includegraphics[width=0.74\textwidth]{images/Train_clip_0_xy.pdf}
\begin{picture}(450,210)
\put(43,-18){\includegraphics[width=0.79\textwidth]{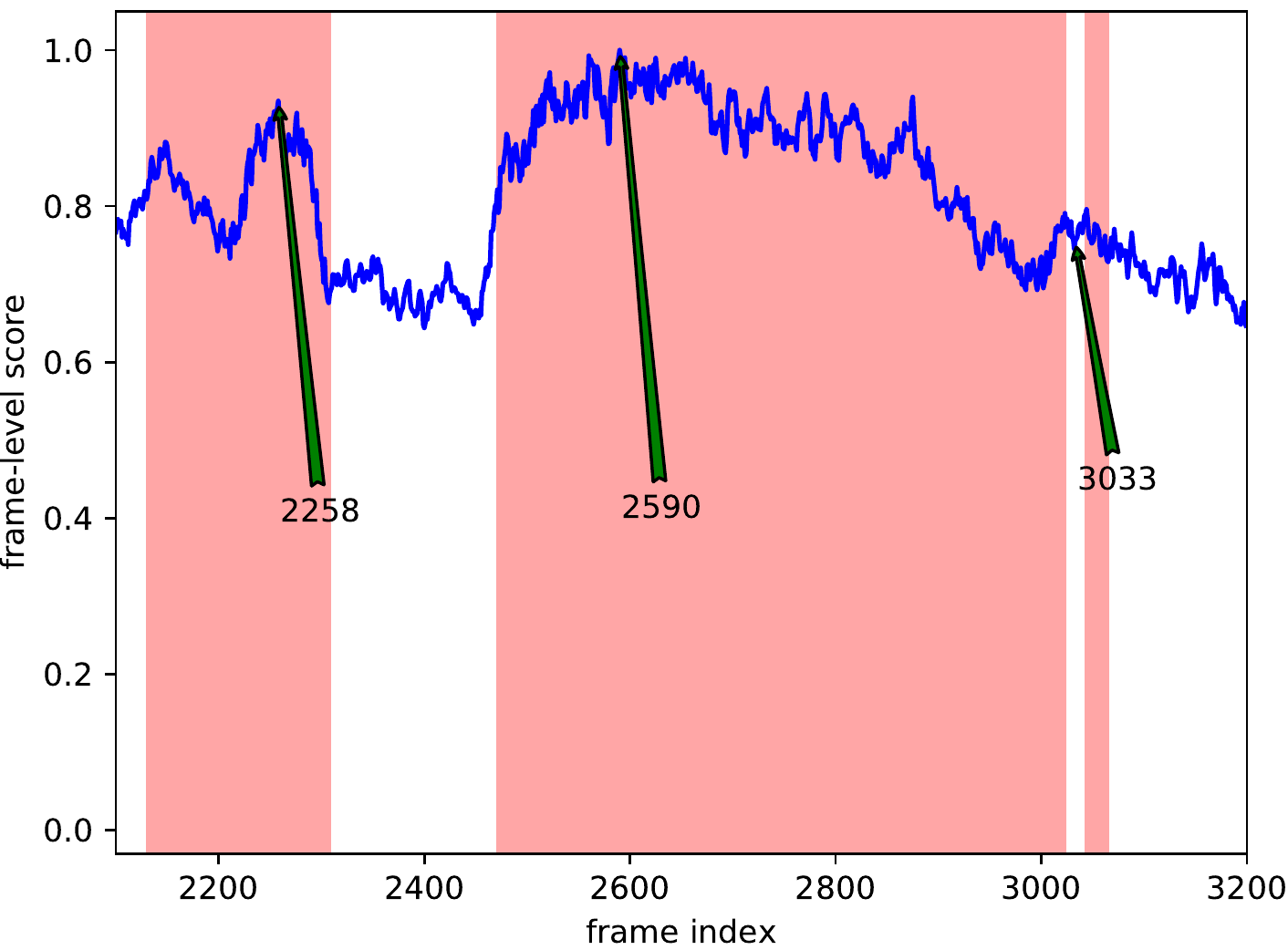}}
\put(80,65){\includegraphics[scale=0.32]{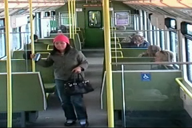}}
\put(160,78){\includegraphics[scale=0.32]{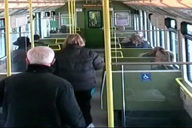}}
\put(259,65){\includegraphics[scale=0.32]{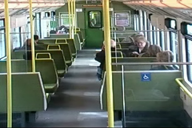}}
\end{picture}
\caption{Frame-scores estimated on Traffic-Train dataset using $\mathcal{S}_{x,y}$. Best viewed in color.}
\label{fig:scores_Train}
\end{figure}

Figs.~\ref{fig:scores_Avenue},~\ref{fig:scores_Ped2},~\ref{fig:scores_Belleview},~\ref{fig:scores_Train} respectively demonstrate the sequences of frame-level scores estimated by our hybrid network (with decoder) on some clips in the Avenue, Ped2, Belleview and Traffic-Train datasets. These results were obtained according to their best score estimation, i.e. using $\mathcal{S}_{R,x,y}$ for the first two datasets, and $\mathcal{S}_{x,y}$ for the remaining.

The red regions indicate anomalous frames that are annotated in the ground truth. The readers are also encouraged to watch the attached video to get a visual understanding.

\bibliography{egbib}